\documentclass[journal]{IEEEtran}
 
\usepackage[]{graphicx}
\graphicspath{{Figures/}}
\usepackage{caption}
\usepackage{subfig} 
\usepackage{amsmath}
\usepackage{amssymb}
\usepackage{epsfig}
\usepackage{cite}
\usepackage{color}
\usepackage{balance}
\usepackage{amsmath}
\usepackage{amsfonts} 
\usepackage{graphicx}
\usepackage{pdfpages}
\usepackage[T1]{fontenc} 
\usepackage{array}
\usepackage{url}

\usepackage{color}
\usepackage{wrapfig}
\usepackage{lipsum}
\usepackage[hidelinks]{hyperref}
\usepackage{multirow}
\usepackage{tabularx}
\usepackage{tikz}

\usepackage{algpseudocode} 
\usepackage{algorithm,algpseudocode}

\title{Rough Set Based Color Channel Selection}

\author{Soumyabrata~Dev,~\IEEEmembership{Student Member,~IEEE,}
Florian~M.~Savoy,~\IEEEmembership{Associate Member,~IEEE,}
Yee~Hui~Lee,~\IEEEmembership{Senior~Member, IEEE,}
and~Stefan~Winkler,~\IEEEmembership{Senior Member,~IEEE}
\thanks{Manuscript received 01-Jul-2016; revised 03-Oct-2016; accepted 28-Oct-2016.}        
\thanks{S.\ Dev and Y.\ H.\ Lee are with the School of Electrical and Electronic Engineering, Nanyang Technological University, Singapore (e-mail: soumyabr001@e.ntu.edu.sg, EYHLee@ntu.edu.sg). }
\thanks{F.\ M.\ Savoy and S.\ Winkler are with the Advanced Digital Sciences Center (ADSC), University of Illinois at Urbana-Champaign, Singapore (e-mail: f.savoy@adsc.com.sg, Stefan.Winkler@adsc.com.sg).}
\thanks{Send correspondence to S.\ Winkler, E-mail: Stefan.Winkler@adsc.com.sg.}
}

\begin{document}
 	
\maketitle
	
\begin{abstract}
Color channel selection is essential for accurate segmentation of sky and clouds in images obtained from ground-based sky cameras. Most prior works in cloud segmentation use threshold based methods on color channels selected in an ad-hoc manner. In this letter, we propose the use of rough sets for color channel selection in visible-light images. Our proposed approach assesses color channels with respect to their contribution for segmentation, and identifies the most effective ones.
\end{abstract}
	
	
\section{Introduction}
\label{sec:intro}
\IEEEPARstart{G}{round-based} whole sky imagers (WSIs) are becoming popular among the remote sensing community. They provide instantaneous data of cloud formations and are thus useful in a variety of applications~\cite{GRSM2016}. WSIs complement satellite images with localized data of higher temporal and spatial resolution. They capture images of the sky at regular intervals and archive them for processing. Most WSIs use traditional cameras in the visible-light spectrum \cite{IGARSS2015}, while a few models capture the near-infrared range as well \cite{infrared_UK,WAHRSIS}. 

Typical post-processing algorithms include the computation of the fraction of the sky covered by clouds~\cite{thincloud,JSTARS2016}, the recognition of cloud types~\cite{ICIP2015b}, or the estimation of the cloud base height~\cite{IGARSS2015b}. A prerequisite for these applications is the segmentation of clouds from the sky, with each pixel of the image classified as either \emph{sky} or \emph{cloud}. As the sky is predominantly blue because of Rayleigh scattering, most existing approaches use thresholding on an ad-hoc combination of red and blue color channels~\cite{Kreuter2009,Li2011}. Kreuter et al.\ \cite{Kreuter2009} used a fixed threshold for the blue/red ratio. Calb{\'o} and Sabburg \cite{Calbo2008} use various statistical features (mean, standard deviation, entropy, etc.) obtained from red and blue channels for successful detection and subsequent labeling of pixels. The difference of red and blue channels is exploited in \cite{Heinle2010,LiuSP2015}. The saturation channel of the HSV color model is used in \cite{Souza}. Recently, Li et al.\ \cite{Li2011} proposed the use of a normalized red/blue ratio for cloud detection. 

We believe only a structured review of color channels allows us to systematically select the one(s) with the most \emph{discriminative} cues. This helps to  efficiently represent the images in a lower-dimensional subspace. Such systematic analysis of color channels for sky/cloud images is important for subsequent tasks, such as cloud type recognition or feature matching in cloud base height estimation.
An analysis of several color channels for sky/cloud segmentation is provided in our previous works, where we used bimodality~\cite{ICIP1_2014} and principal component analysis~\cite{JSTARS2016} to determine favorable channels. However, these techniques had certain shortcomings with regards to ranking color channels according to their relevance for cloud segmentation, as their correlation with cloud segmentation accuracy is relatively weak. Several other techniques have been  used for selecting the most discriminative features in a classification problem. Serrano et al.\ \cite{fs-ROC} chose the area under the Receiver Operator Characteristic (ROC) curves \cite{Kerekes08} for feature selection in a synthetic dataset. Information-theoretic measures like Kullback-Leibler (KL) distance or divergence have been applied to band selection in hyperspectral imaging~\cite{InfoTheory_TGRS}. 

Our main contribution in this letter is to determine those color channels that are most discriminative in identifying cloud pixels in traditional visible-light images. We extend the benchmarking done in \cite{ICIP1_2014} using a \emph{rough set} based approach that can accurately assess the efficiency of different color channels for cloud segmentation.\footnote{~The source code of all simulations in this paper is available online at \url{https://github.com/Soumyabrata/rough-sets}.}
Rough set theory, originally introduced by Pawlak~\cite{Pawlak92}, is useful for representing uncertain data with a level of approximation, and selecting the most discriminative features from the feature space. Recently, it has been successfully applied to hyperspectral band selection~\cite{rs-TGRS}. To the best of our knowledge, our proposed approach is the first that uses rough set theory for color channel selection in visible-light images. 
	
The remaining letter is organized as follows. In Section~\ref{sec:rs}, we describe the fundamental concepts behind rough set theory and present our proposed color channel selection algorithm. Experimental results are presented in Section~\ref{sec:exp}. Finally, Section~\ref{sec:conclusion} concludes the letter.

\section{Rough Set Based Color Channel Selection}
\label{sec:rs}
Classical rough sets \cite{Pawlak92} are an approximation of conventional sets in set theory. In a scenario where it is difficult to define the boundaries of a conventional set, rough set theory provides a set of mathematical tools to  define them in a approximate way. It facilitates an objective analysis in a data-driven system which is vague, uncertain and incomplete. Here we first explain the related terminologies of rough sets, and subsequently define the use of rough sets for visible-light images.

\subsection{Rough Set Theory}
In rough sets, information is expressed in the form of a \emph{decision table}. We define a decision table $\mathcal{L}$ such that each row represents an observation, and each column is an attribute from attribute set  $\mathcal{A}$. This non-empty set of observations is usually referred to as the \emph{universe} $\mathcal{U}$.
Formally, for each entry in the decision table, we define the function $f$ that maps attribute $\mathcal{A}$ to value domain $\mathcal{V}$, $f: \mathcal{U} \times \mathcal{A} \rightarrow \mathcal{V}$.
	
Any reduct $\mathcal{P}$ from the set of attributes $\mathcal{A}$ satisfies the indiscernibility (or  equivalence) relation IND($\mathcal{P}$). For any reduct $\mathcal{P} \in \mathcal{A}$, the $\mathcal{P}$-indiscernibility relation is defined as:
\begin{align}
\label{eq:p-ind}
\mbox{IND}(\mathcal{P}) = \{(x_m,x_n) \in \mathcal{U}^2 | \forall a \in \mathcal{P}, f(x_m,a) = f(x_n,a)\},
\end{align}
where $x_m$ and $x_n$ are two observations from the universe $\mathcal{U}$, and $a$ is an element from set $\mathcal{P}$. This indicates that $x_m$ and $x_n$ are indiscernable based on the attribute $\mathcal{P}$, as the value function $f$ assigns both $x_m$ and $x_n$ to the same value.
This partition of $\mathcal{U}$ generated by IND($\mathcal{P}$) is denoted as:
\begin{align}
\label{eq:p-part}
\mathcal{U}/\mbox{IND}(\mathcal{P}) = \{[x_m]_{\mathcal{P}} | x_m \in \mathcal{U}\}.
\end{align}
	
Let $\mathcal{X}$ be a set of observations from the universe $\mathcal{U}$. Rough set theory asks the question: how can we express this conventional set $\mathcal{X}$, using only the information in attribute set $\mathcal{P}$? In general, there is no precise answer, and therefore approximations are generated. They are defined by their corresponding $\mathcal{P}$-\emph{lower}- and $\mathcal{P}$-\emph{upper}-approximations as:

\begin{subequations}
\label{eq:low-upp}
\begin{equation}
\underline{\mathcal{P}}(\mathcal{X}) = \cup \{[x_m] | [x_m] \subseteq \mathcal{X}\},
\end{equation}    
\begin{equation}
\overline{\mathcal{P}}(\mathcal{X}) = \cup \{[x_m] | [x_m] \cap \mathcal{X} \neq \phi\}.
\end{equation}
\end{subequations}
	
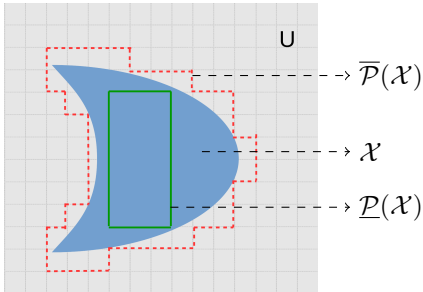
\begin{figure}[htb]
\centering
\begin{tikzpicture}
\begin{scope}[xshift=1.5cm]
    \node[anchor=south west,inner sep=0] (image) at (0,0) {\includegraphics[width=0.23\textwidth]{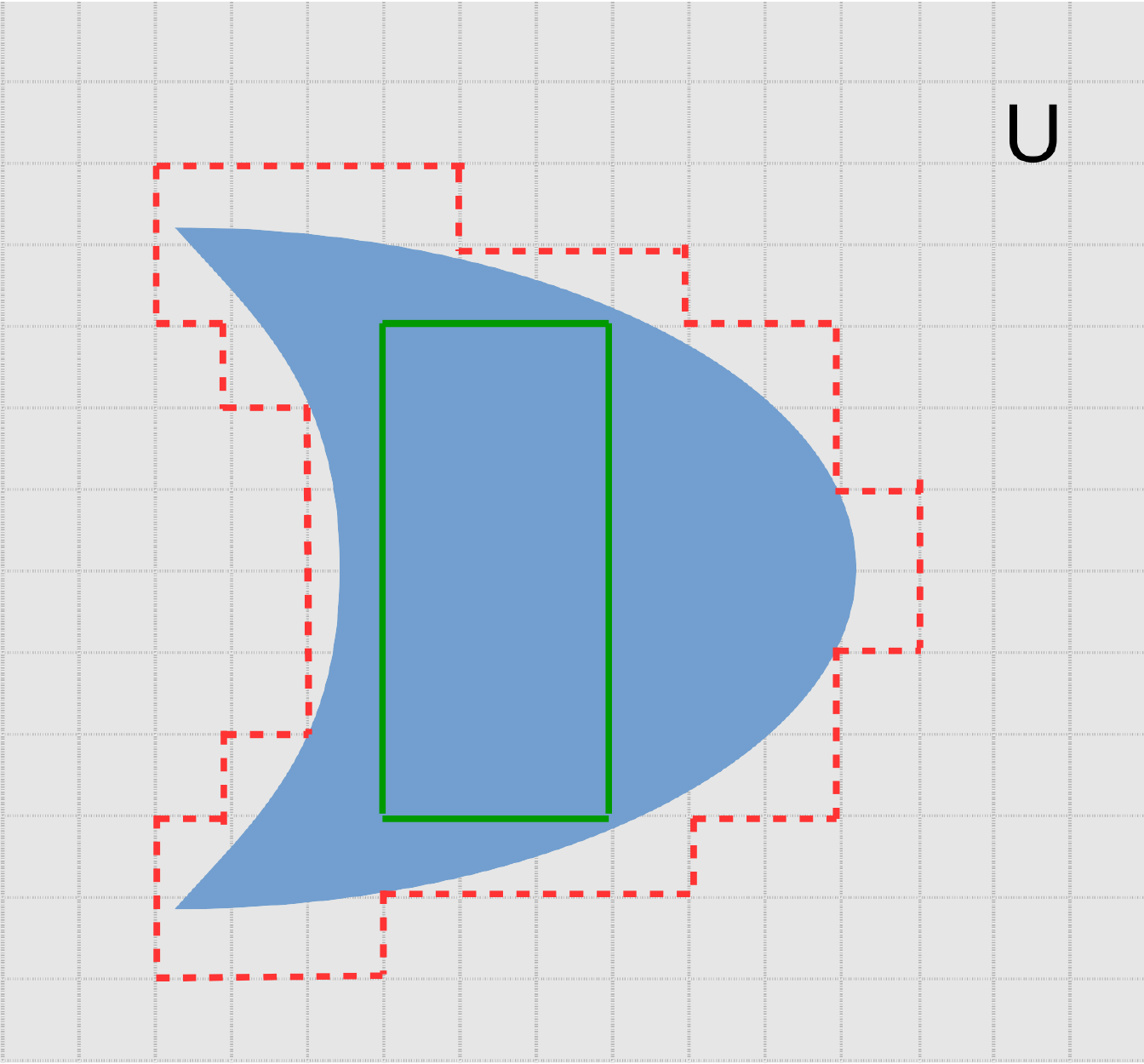}};
    \begin{scope}[x={(image.south east)},y={(image.north west)}]
        \draw[->,black,dashed](0.6,0.75) -- (1.1,0.75) node[anchor=west,black] {$\overline{\mathcal{P}}(\mathcal{X})$};
        \draw[->,black,dashed](0.63,0.49) -- (1.1,0.49) node[anchor=west,black] {$\mathcal{X}$};
        \draw[->,black,dashed](0.53,0.3) -- (1.1,0.3) node[anchor=west,black] {$\underline{\mathcal{P}}(\mathcal{X})$};        
    \end{scope}
\end{scope}
\end{tikzpicture}
\caption{Illustration of a typical rough set, which approximates a conventional set $\mathcal{X}$ (depicted in blue). Each individual grid depicts a partition of the universe generated by an equivalence relation. The union of all such partitions indicated with solid green borders (definite members) represents the \emph{lower approximation} $\underline{\mathcal{P}}(\mathcal{X})$; while the dotted red borders (possible members) represent the \emph{upper approximation} $\overline{\mathcal{P}}(\mathcal{X})$.}
\label{fig:rs-illus}
\end{figure}

The observations in the lower approximation set $\underline{\mathcal{P}}(\mathcal{X})$ are the definite members of $\mathcal{X}$, also called the \emph{positive} region $\mbox{POS}(\mathcal{X})$. On the other hand, $\overline{\mathcal{P}}(\mathcal{X})$ represents the upper approximation of the set. It denotes the possible members of $\mathcal{X}$, based on the knowledge in the decision table. 
We illustrate this in Fig.\ref{fig:rs-illus}.
	
Let us assume that the attribute set $\mathcal{A}$ consists of both condition and decision attributes $\mathcal{C}$ and $\mathcal{D}$ respectively, such that $\mathcal{A} = \mathcal{C} \cup \mathcal{D}$. The \emph{relevance} criterion is defined as the dependence between $\mathcal{C}$ and $\mathcal{D}$, and can be expressed as:
\begin{align}
\label{eq:rel-defn}
\gamma_\mathcal{C}(\mathcal{D}) = \frac{|\mbox{POS}_\mathcal{C}(\mathcal{D})|}{|\mathcal{U}|},
\end{align}
where |$\cdot$| denotes the cardinality of the set. This dependence value is an important measure to identify the most discriminate attribute from set $\mathcal{A}$. The value of $\gamma_\mathcal{C}(\mathcal{D})$ ranges between $0$ and $1$, where $0$ indicates independence and $1$ indicates $\mathcal{D}$ fully depends on $\mathcal{C}$.
	
\subsection{Color Channel Selection Using Rough Sets}

\begin{figure*}[htb]
\centering
\makebox[0.1\textwidth][c]{\hspace{-1mm}\small{Input Image}}
\makebox[0.1\textwidth][c]{$c_1$}
\makebox[0.1\textwidth][c]{$c_2$}
\makebox[0.1\textwidth][c]{$c_3$}
\makebox[0.1\textwidth][c]{$c_4$}
\makebox[0.1\textwidth][c]{$c_5$}
\makebox[0.1\textwidth][c]{$c_6$}
\makebox[0.1\textwidth][c]{$c_7$}
\makebox[0.1\textwidth][c]{$c_8$}\\    
\vspace{0.5mm}
\includegraphics[width=0.1\textwidth]{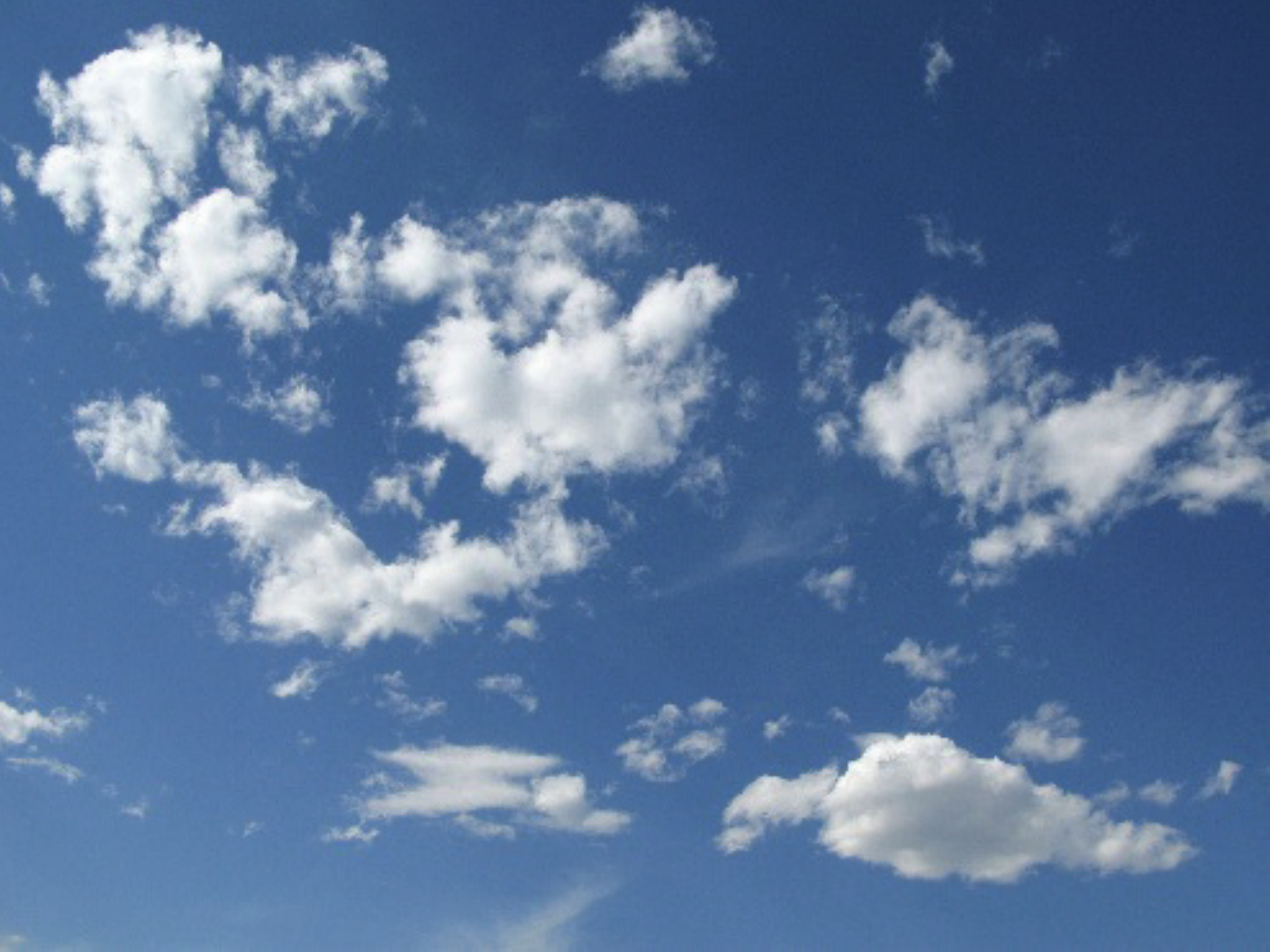}
\includegraphics[width=0.1\textwidth]{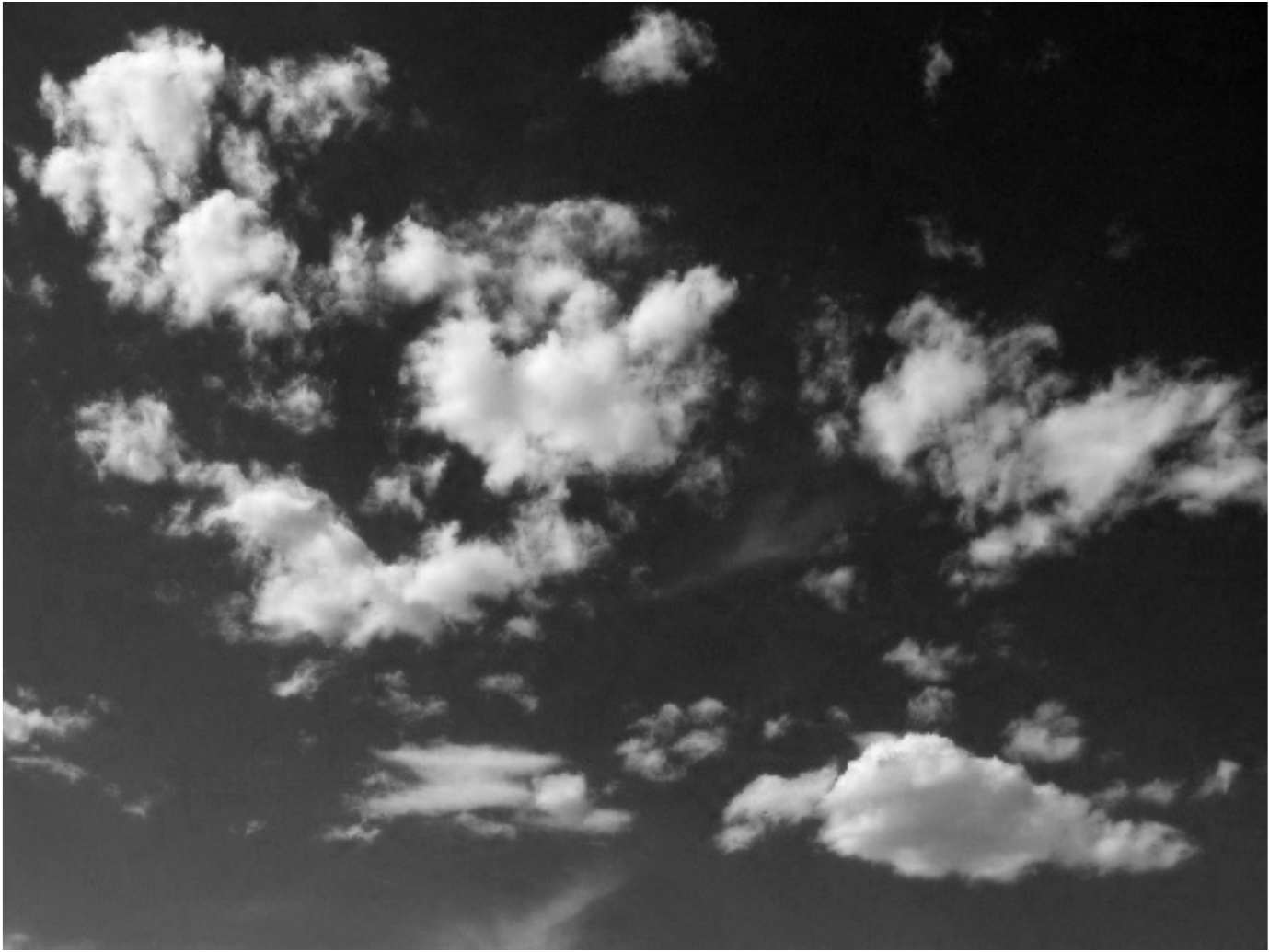}
\includegraphics[width=0.1\textwidth]{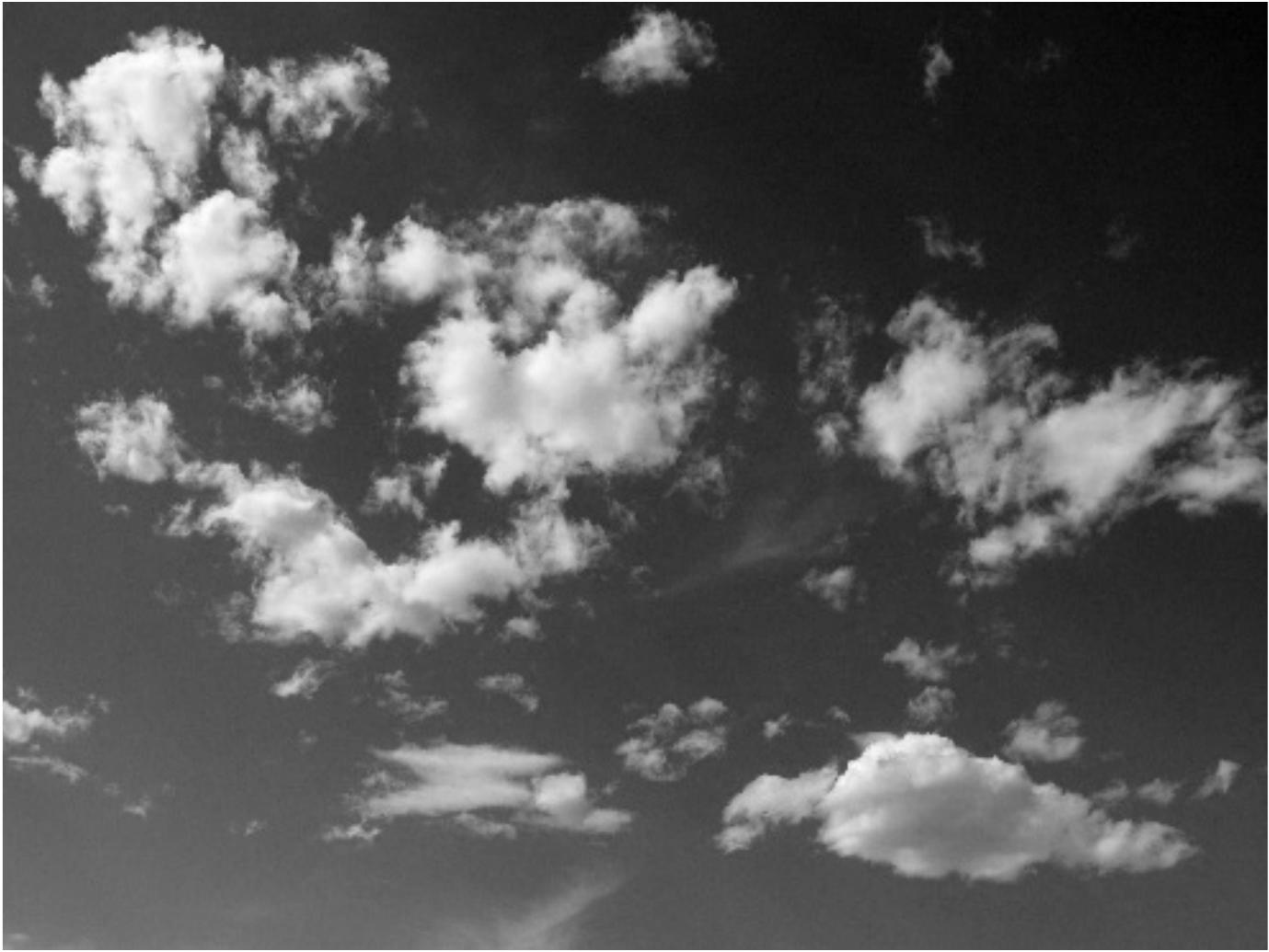} 
\includegraphics[width=0.1\textwidth]{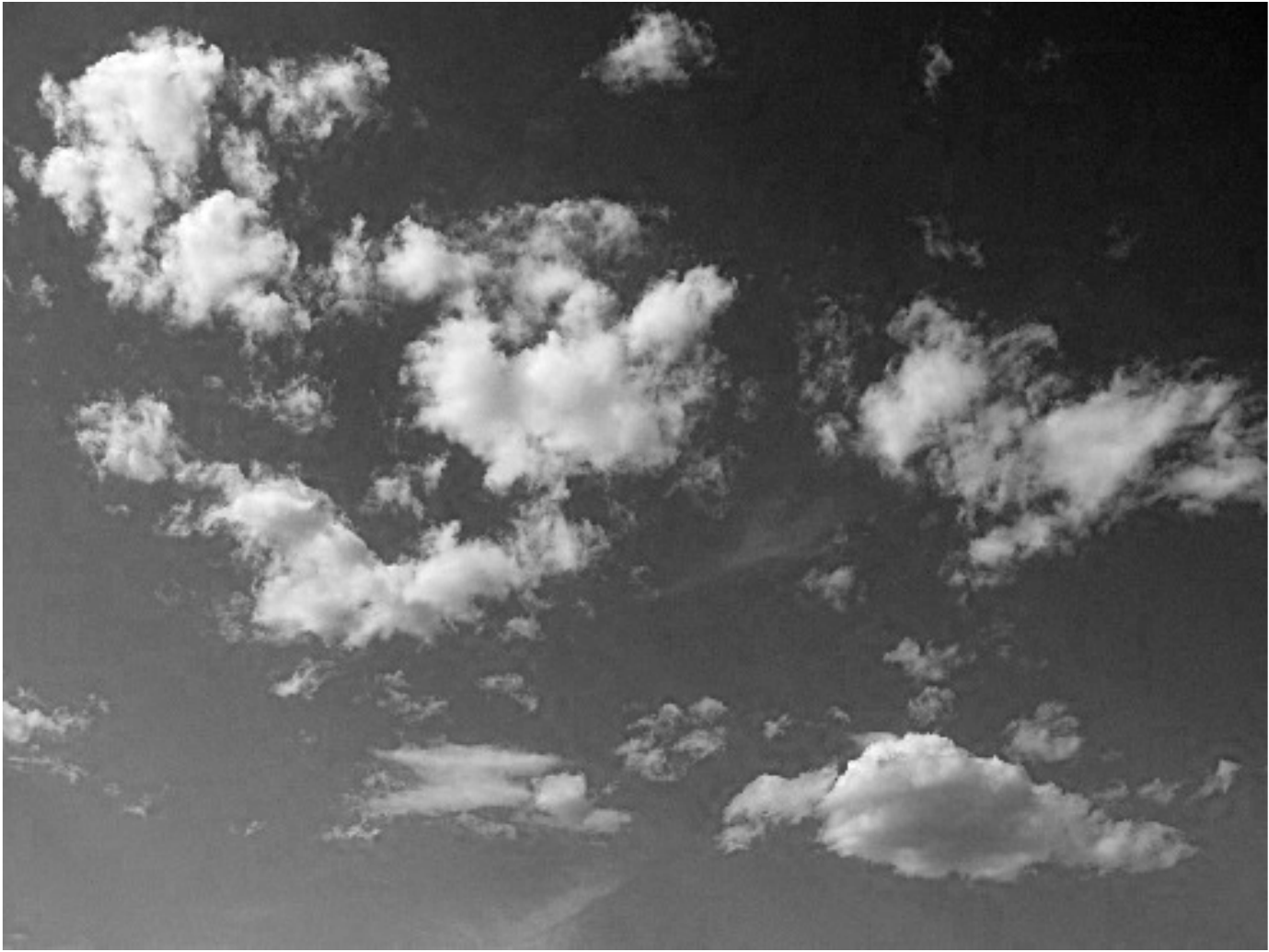}   
\includegraphics[width=0.1\textwidth]{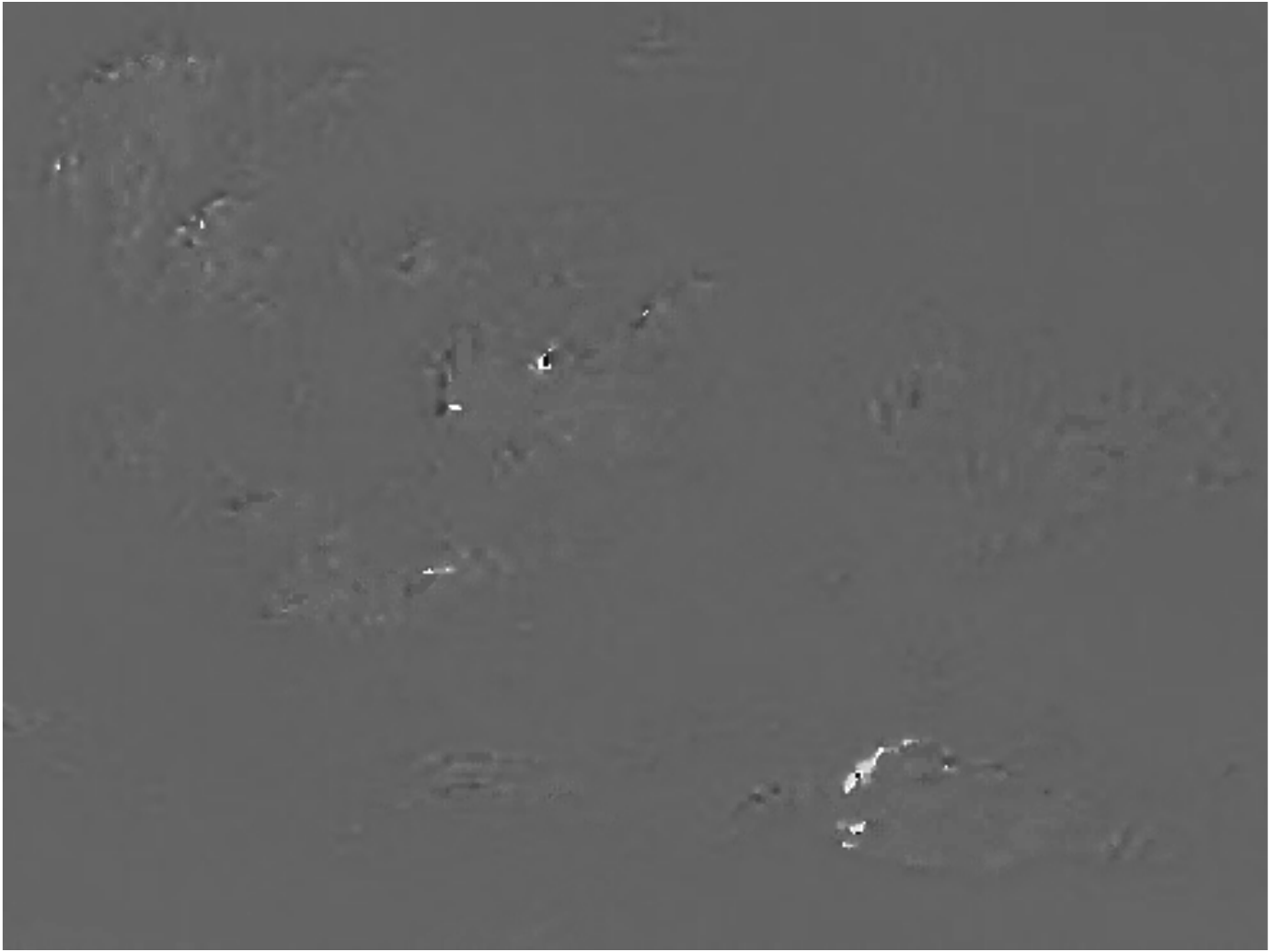}
\includegraphics[width=0.1\textwidth]{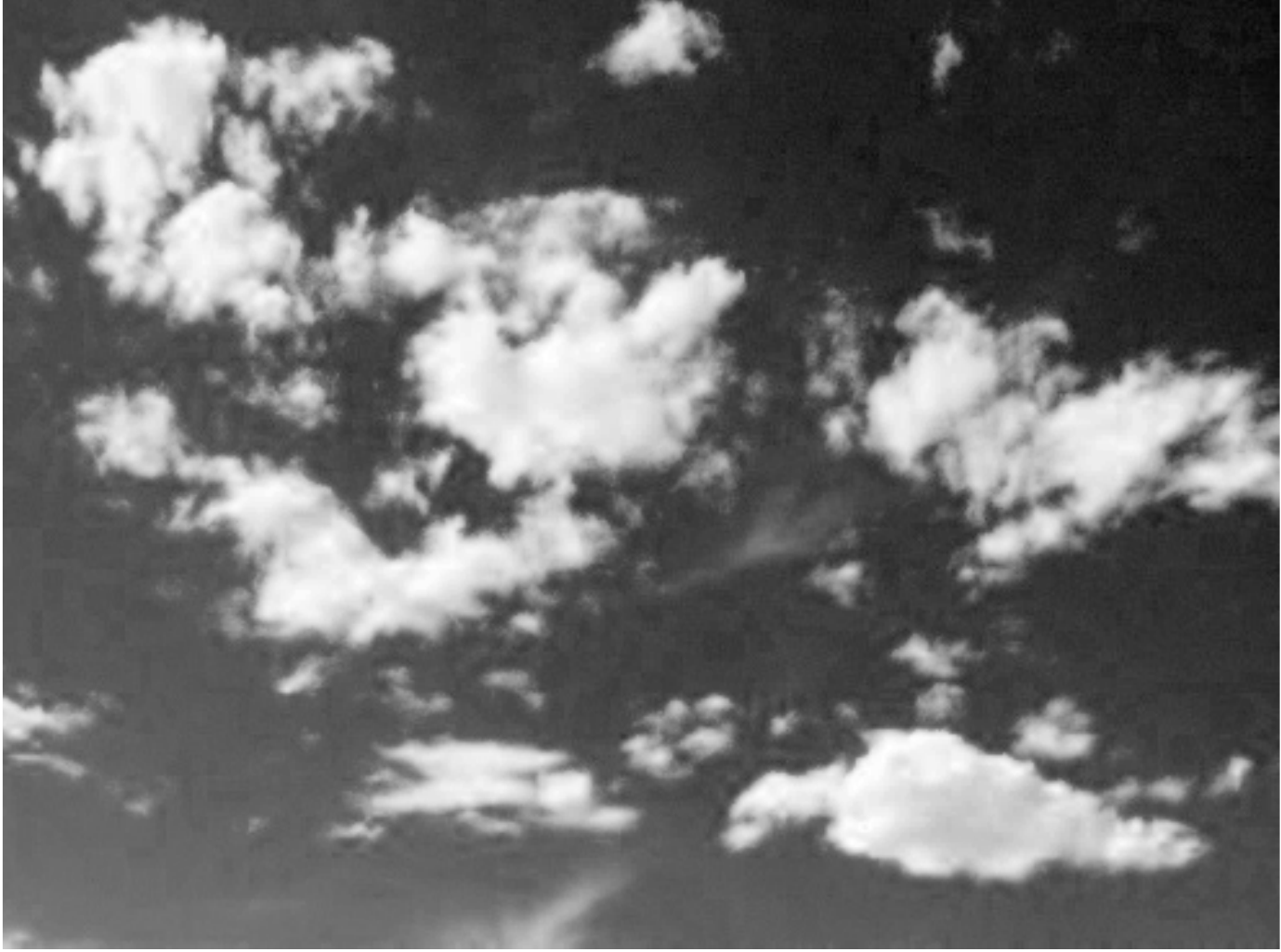}
\includegraphics[width=0.1\textwidth]{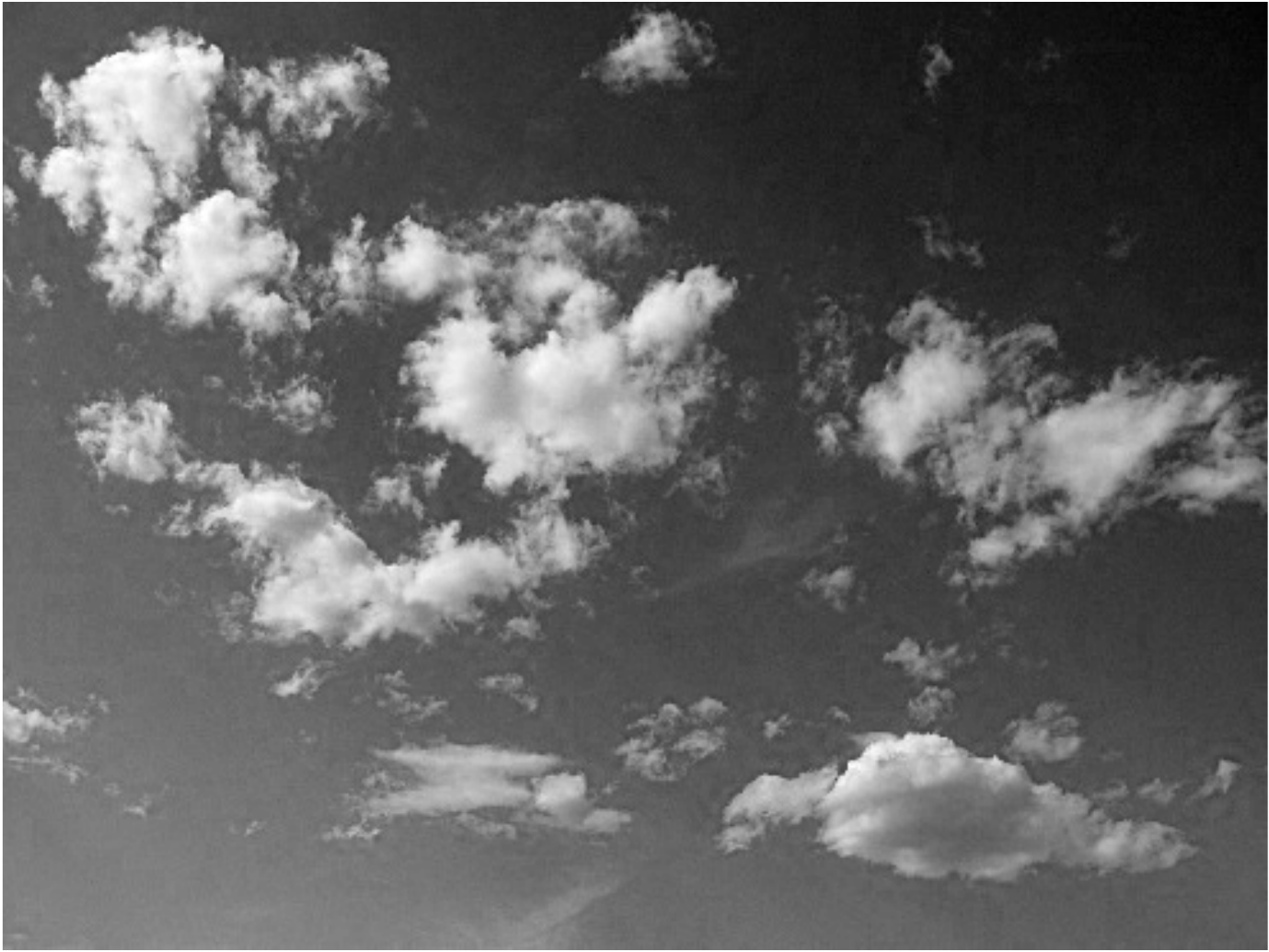}
\includegraphics[width=0.1\textwidth]{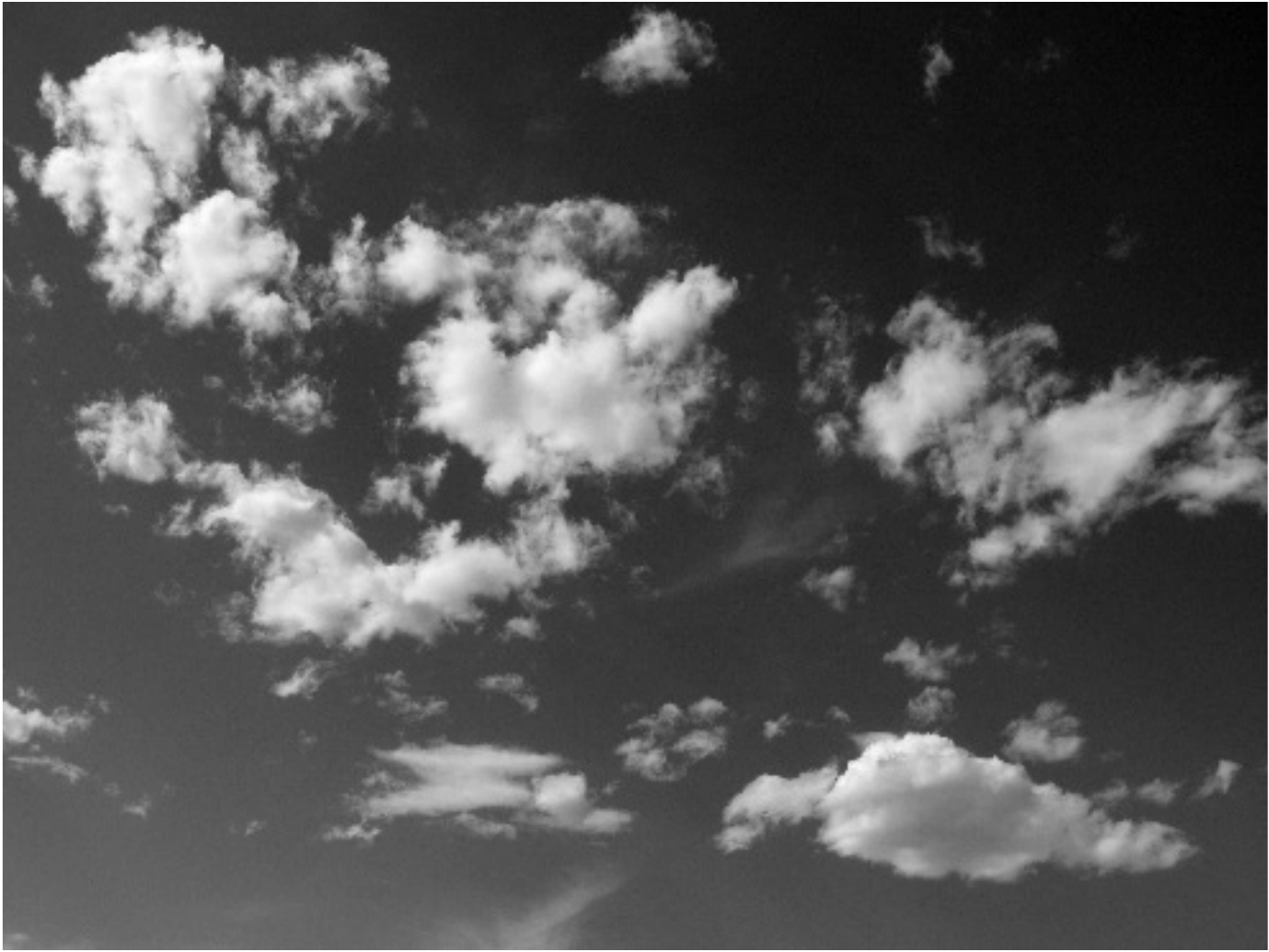}
\includegraphics[width=0.1\textwidth]{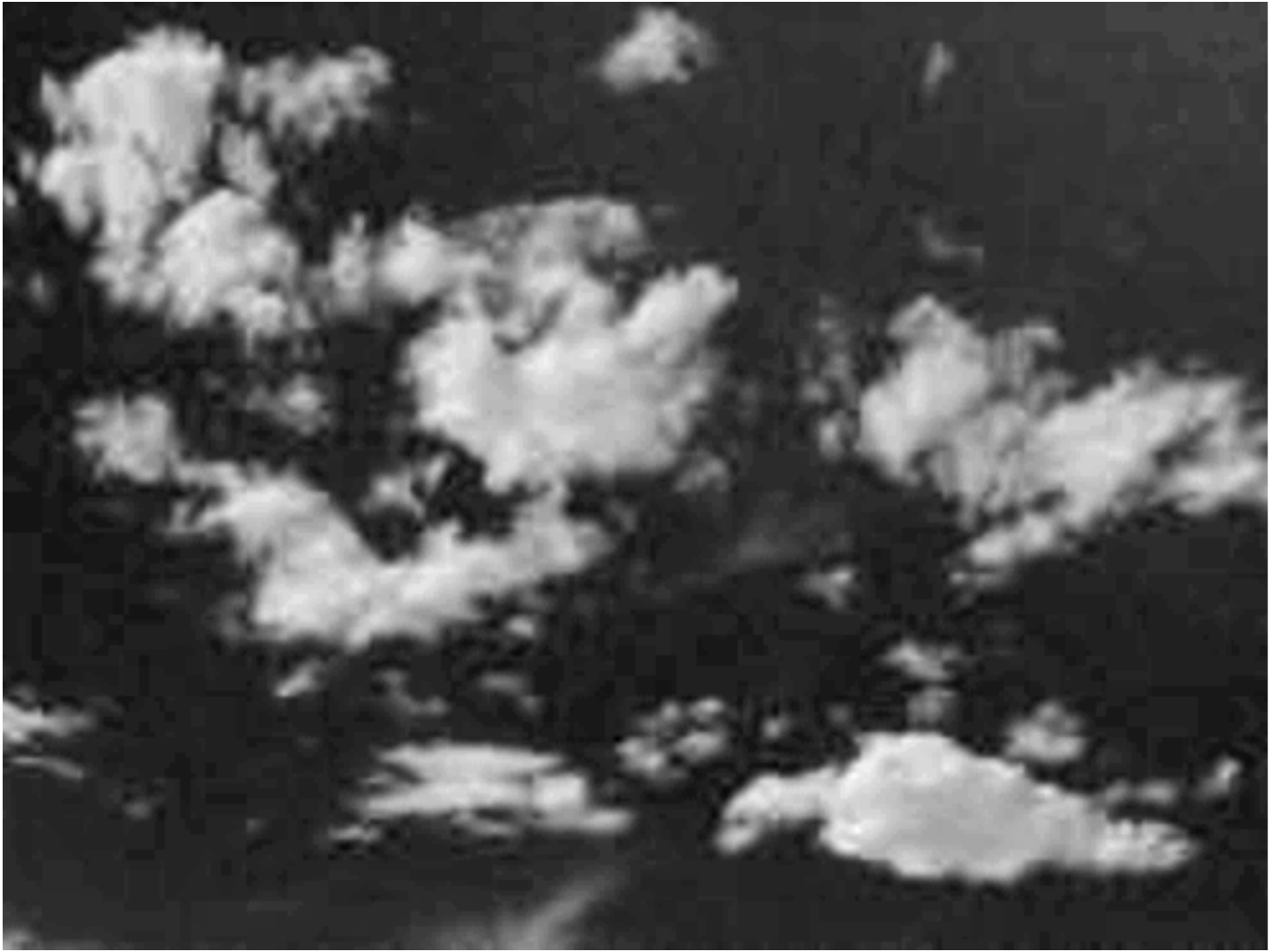}\\
\vspace{1mm}
\includegraphics[width=0.1\textwidth]{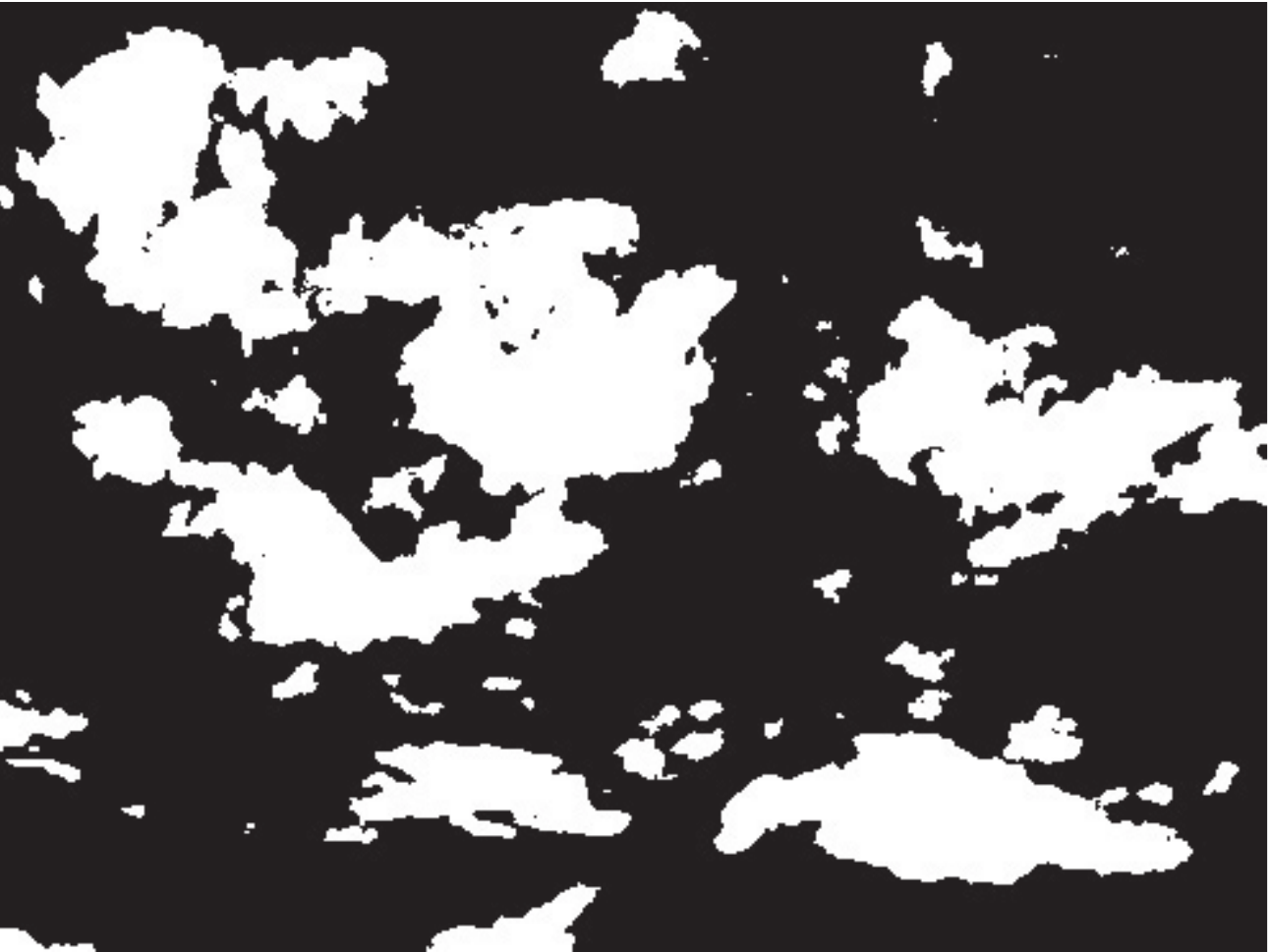}
\includegraphics[width=0.1\textwidth]{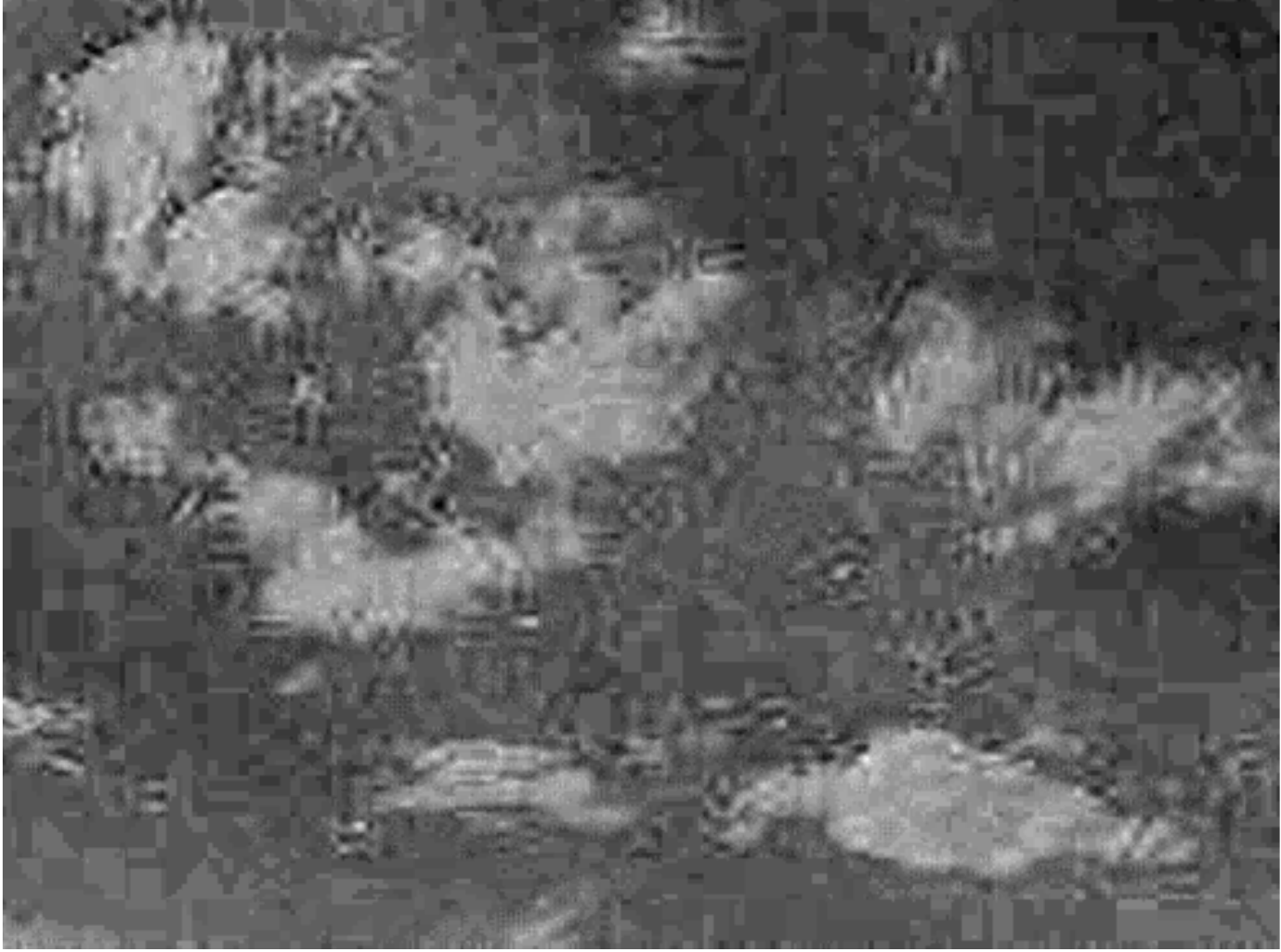}
\includegraphics[width=0.1\textwidth]{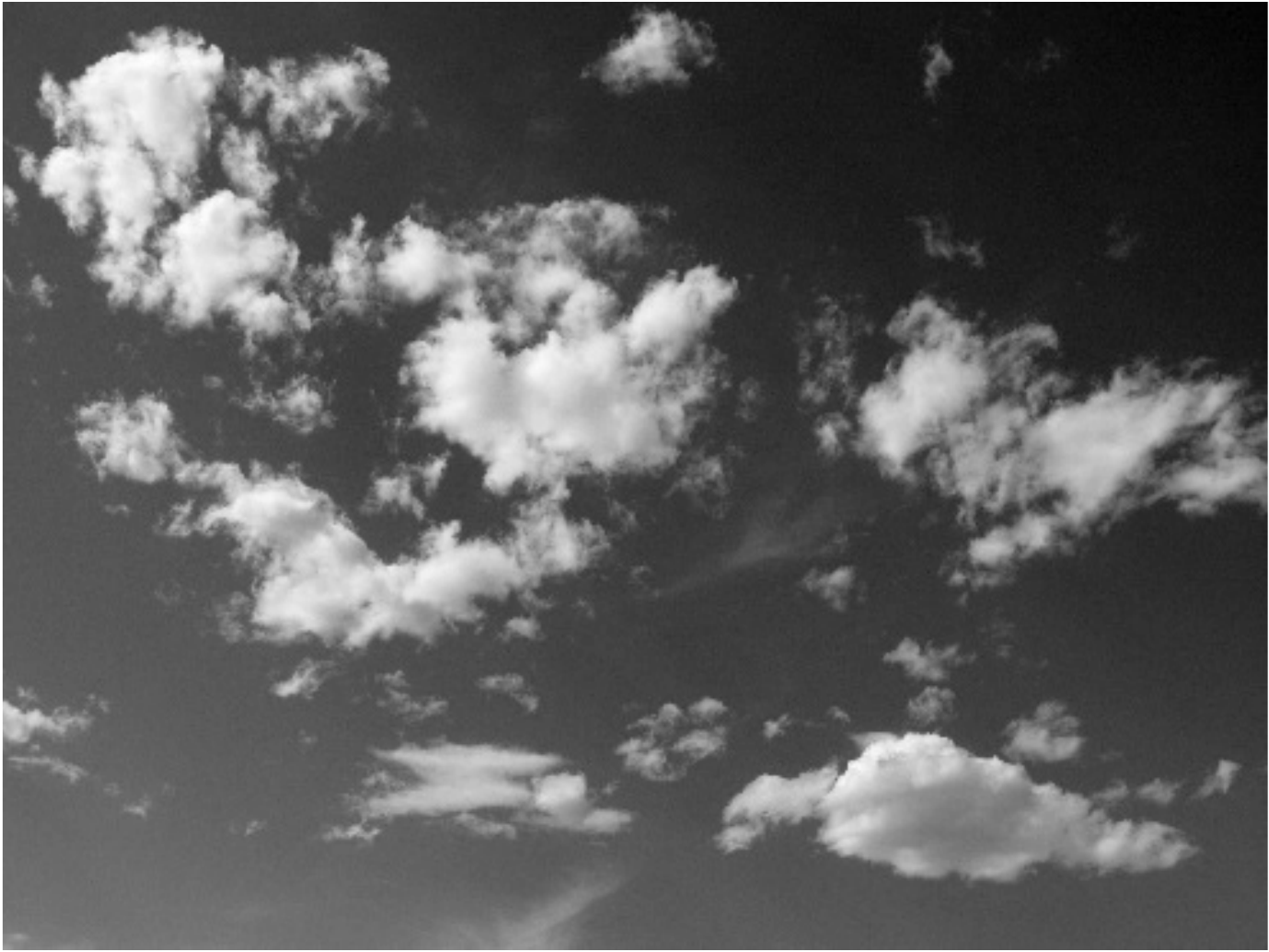}
\includegraphics[width=0.1\textwidth]{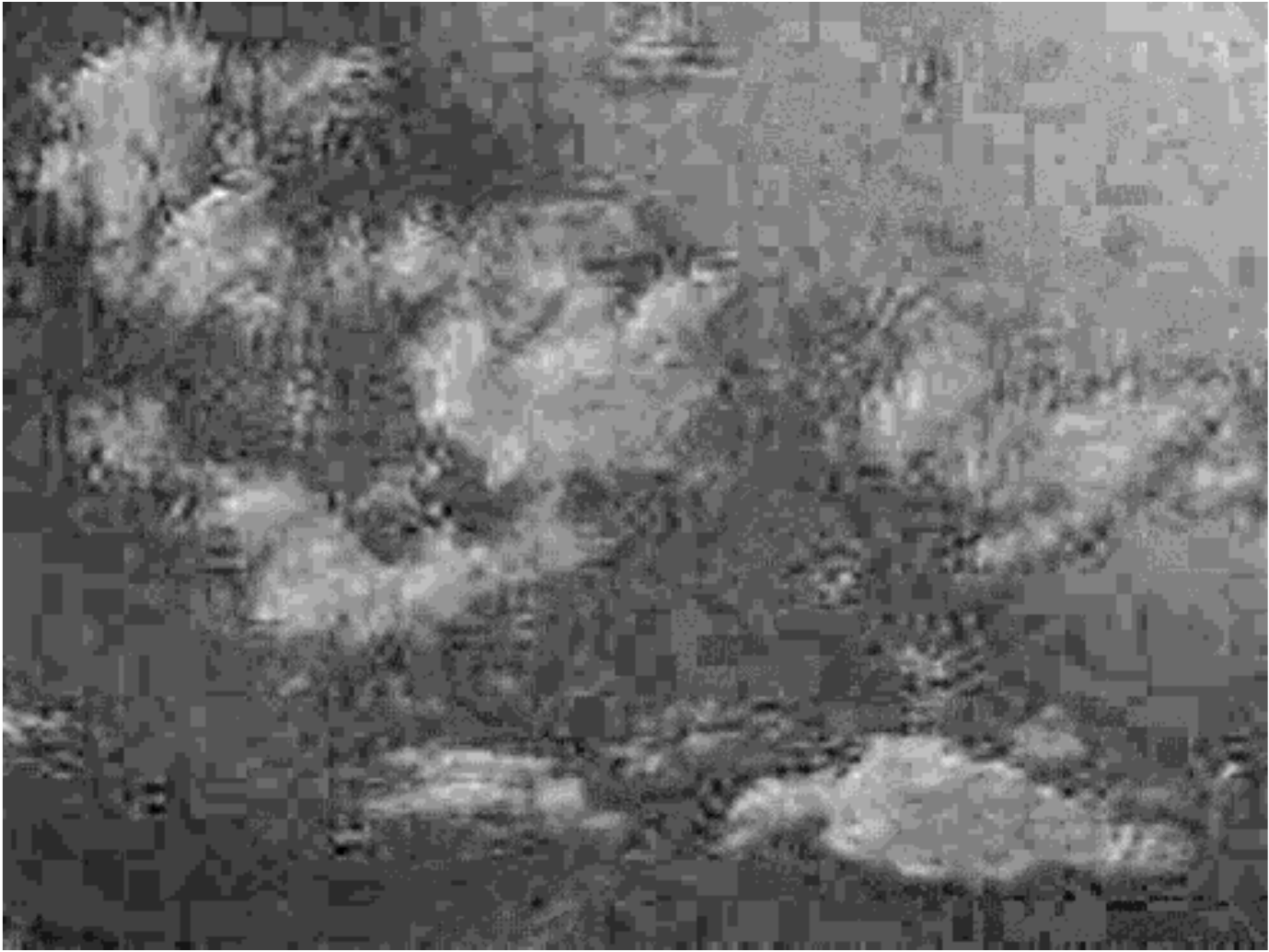} 
\includegraphics[width=0.1\textwidth]{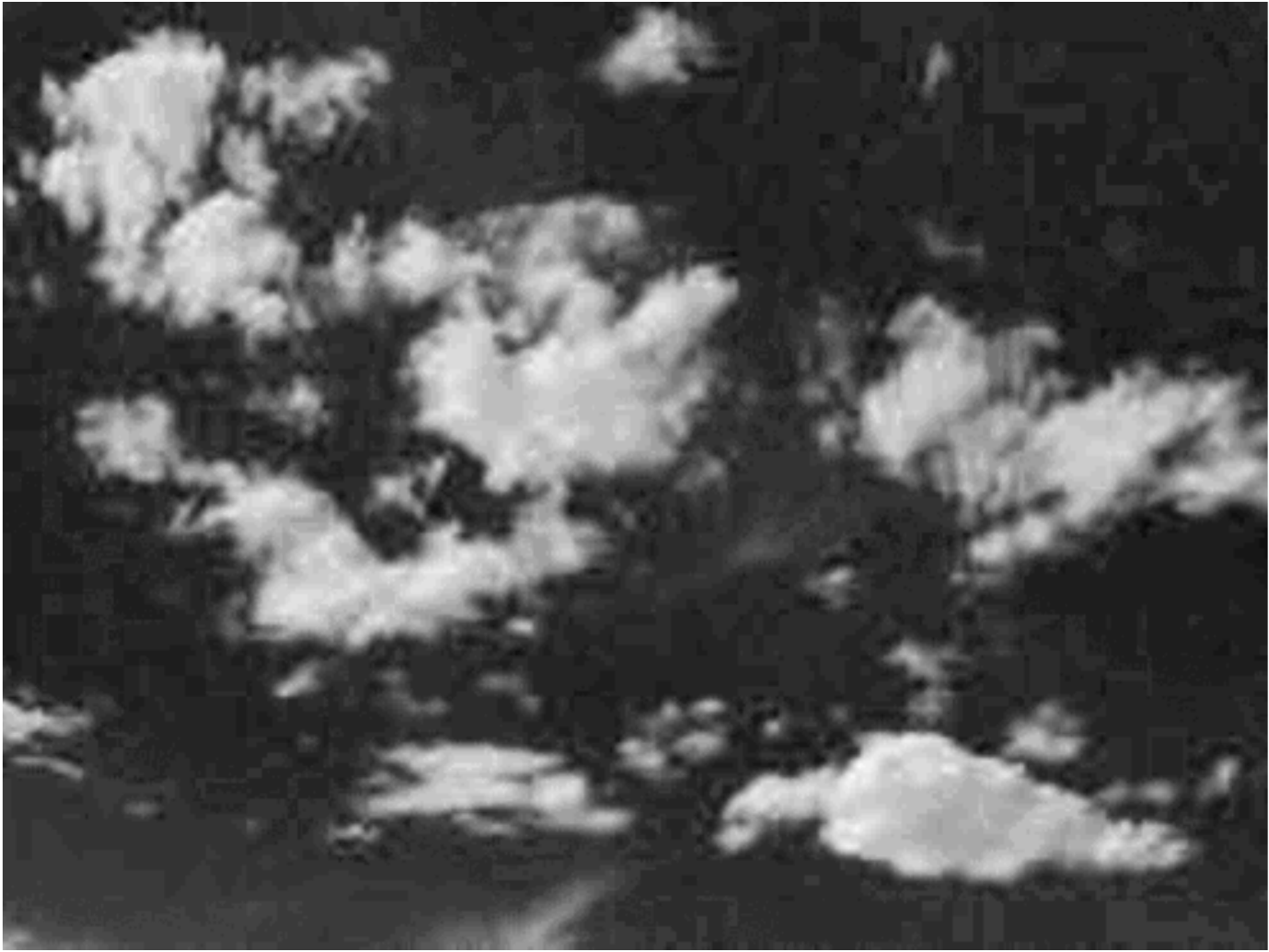}
\includegraphics[width=0.1\textwidth]{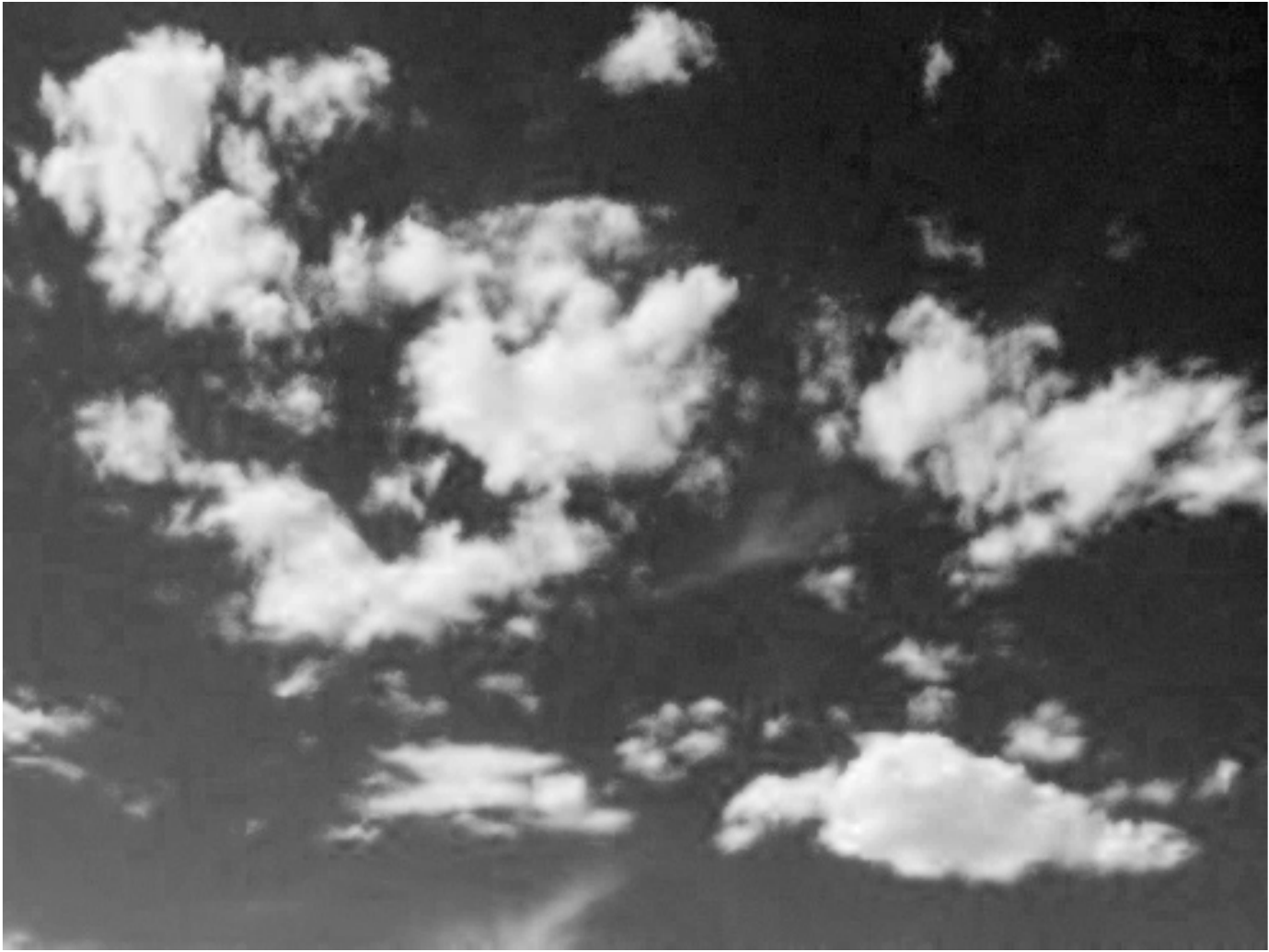}
\includegraphics[width=0.1\textwidth]{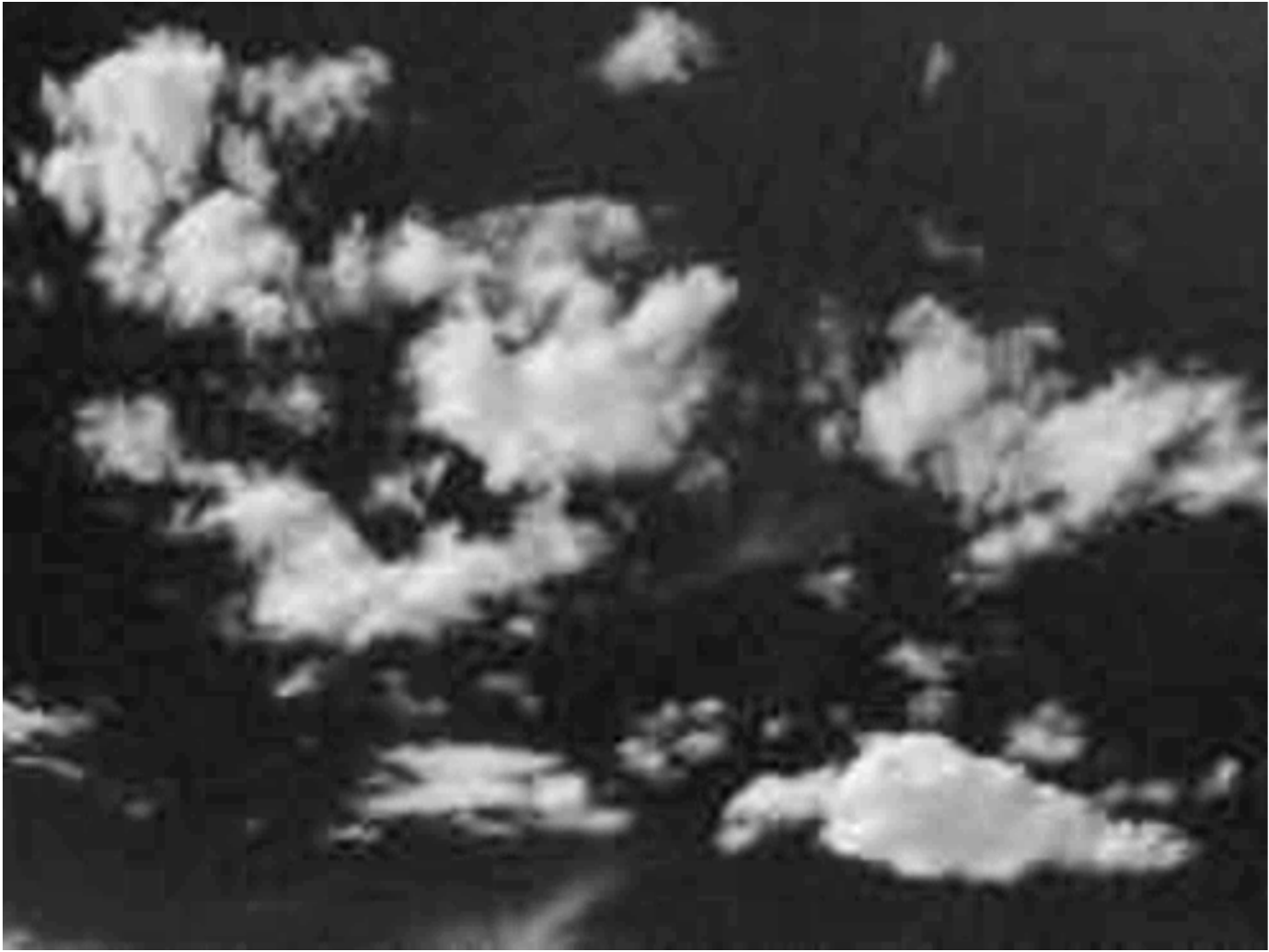}
\includegraphics[width=0.1\textwidth]{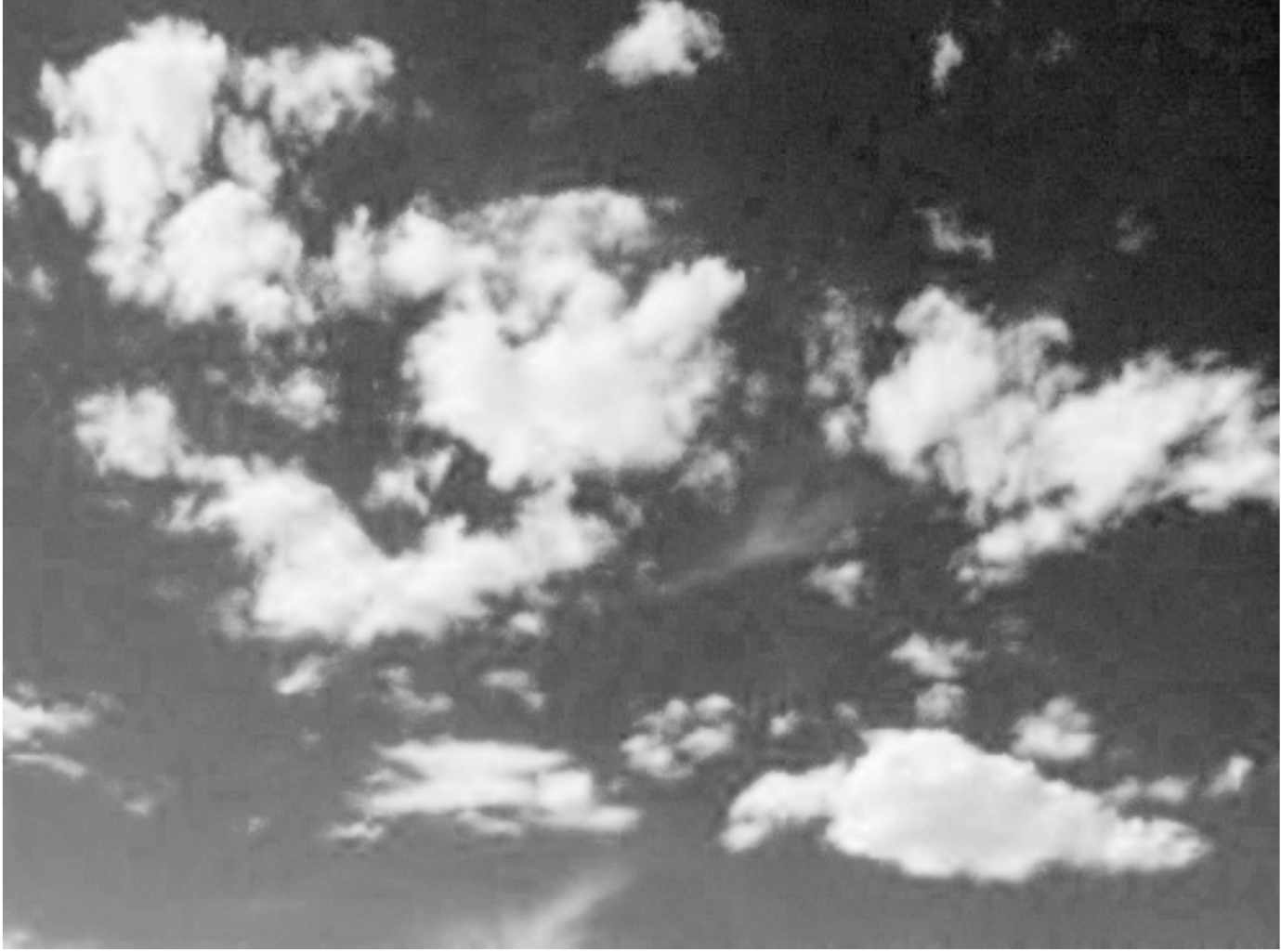}
\includegraphics[width=0.1\textwidth]{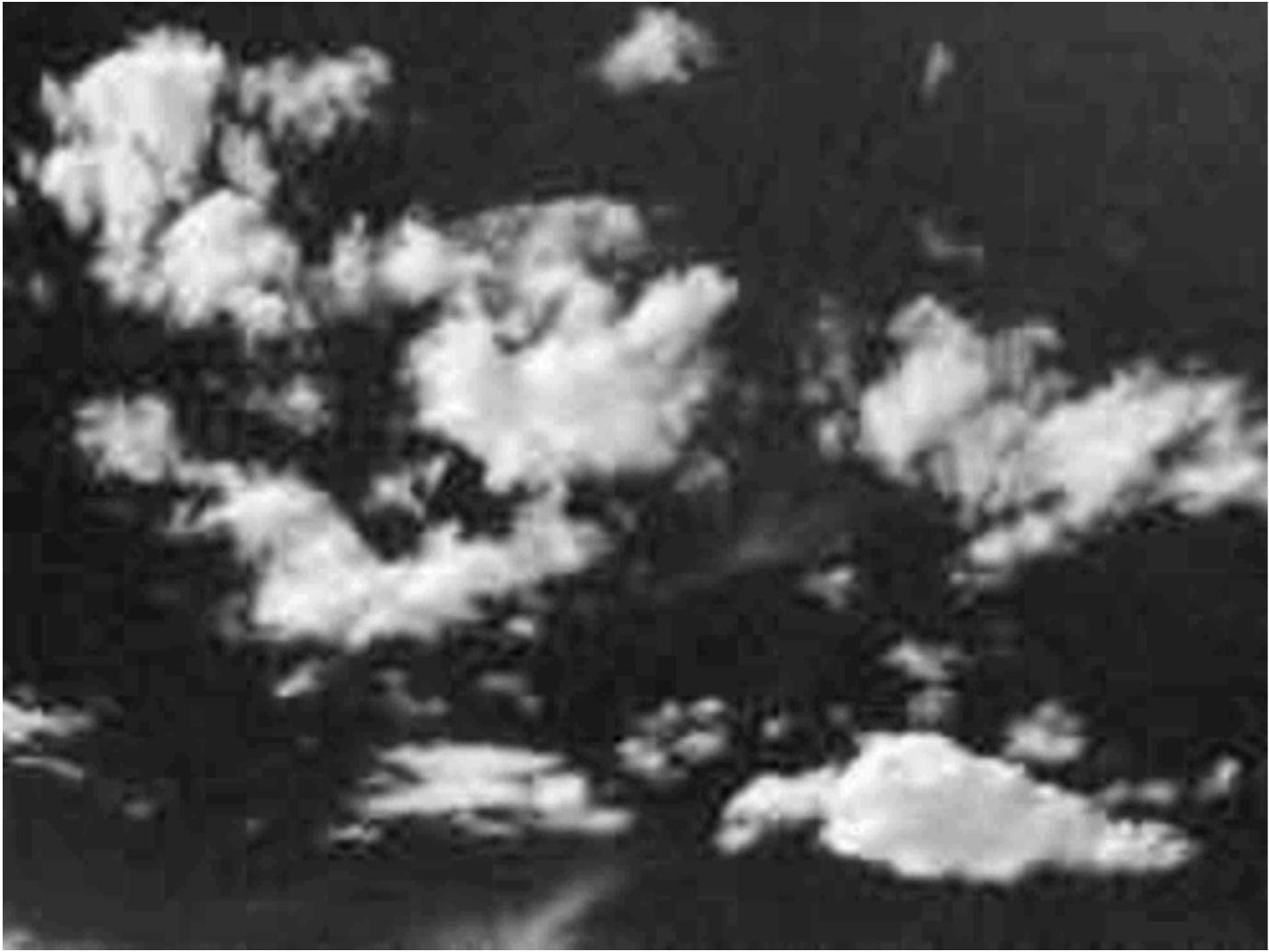}\\
\makebox[0.1\textwidth][c]{\hspace{-2mm}\small{Ground Truth}}
\makebox[0.1\textwidth][c]{$c_9$}
\makebox[0.1\textwidth][c]{$c_{10}$}
\makebox[0.1\textwidth][c]{$c_{11}$}
\makebox[0.1\textwidth][c]{$c_{12}$}
\makebox[0.1\textwidth][c]{$c_{13}$}
\makebox[0.1\textwidth][c]{$c_{14}$}
\makebox[0.1\textwidth][c]{$c_{15}$}
\makebox[0.1\textwidth][c]{$c_{16}$}
\caption{Sample image and segmentation mask from the HYTA database~\cite{Li2011}, together with the $16$ color channels from Table~\ref{tab:color-ch} (the color maps for $c_4$, $c_5$, $c_9$, $c_{15}$, and $c_{16}$ are inverted such that cloud pixels always have a lighter shade than sky pixels).}
\label{fig:objective}
\end{figure*}

We now describe the problem of color channel selection using rough set terminology and propose our  algorithm for this purpose. The main benefit of using rough set theory is that it provides a systematic method to approximate the segmentation ground truth, with the highest degree of approximation. Moreover, no prior information about the data is needed for the analysis.

We consider $16$ color channels in our analysis, as shown in Table~\ref{tab:color-ch} and  illustrated in Fig.~\ref{fig:objective}. They include $RGB$, $HSV$, $YIQ$, CIE $L^{*}a^{*}b^{*}$ color models, various red-blue combinations, and chroma  $C=\max(R,G,B)-\min(R,G,B)$. Several of these color channels are commonly used for thresholding in sky/cloud images \cite{Li2011,Souza,ICIP1_2014}. We utilize rough set theory to identify the most \emph{informative} color channel(s) (a.k.a.\ reducts) from these $16$.

\begin{table}[htb]
\small
\centering
\setlength{\tabcolsep}{4pt} 
\begin{tabular}{c|c||c|c||c|c||c|c||c|c||c|c}
\hline
$c_{1}$ & $R$ & $c_{4}$ & $H$ & $c_{7}$ & $Y$ & $c_{10}$ & $L^{*}$ & $c_{13}$ & $R/B$ & $c_{16}$ & $C$\\
$c_{2}$ & $G$ & $c_{5}$ & $S$ & $c_{8}$ & $I$ & $c_{11}$ & $a^{*}$ & $c_{14}$ & $R-B$& $ $ & $ $\\
$c_{3}$ & $B$ & $c_{6}$ & $V$ & $c_{9}$ & $Q$ & $c_{12}$ & $b^{*}$ & $c_{15}$ & $\frac{B-R}{B+R}$ & $ $ & $ $\\
\hline
\end{tabular}
\caption{Color spaces and components used for analysis.}
\label{tab:color-ch}
\end{table}

Suppose that $\mathcal{U}_i$ is a non-empty finite universe of pixel observations for a single sky/cloud image $\mathbf{I}_i$ from the image dataset $\mathcal{T}=\{\mathbf{I}_1,\mathbf{I}_2, \ldots, \mathbf{I}_N\}$. The set of pixels corresponding to the ground-truth observation for an image $\mathbf{I}_i$ is $\mathcal{G}_i$. We define $\mathcal{A}_i$ as the set of its corresponding vectorized color channels \{$\mathbf{c}_1$, $\mathbf{c}_2$, \ldots, $\mathbf{c}_{16}$\} for a particular image $\mathbf{I}_i$, along with the vectorized decision attribute $\mathbf{g}_i$. These $16$ color channels are called condition attribute sets, and one ground truth vectorized image is the decision attribute set from the family of attributes $\mathcal{A}_i$. 

We define a decision table $\mathcal{L}_i$ such that each row represents a pixel, and each column represents an attribute. 
Thus, for a single image $\mathbf{I}_i$, the function $f$ assigns a value $v_i^{kj}$ in the value domain to each variable-attribute pair ($q_k$,$a_j$), where $q_k$ is the $k$-th pixel value of the image and $a_j$ is the $j$-th attribute from the set $\mathcal{A}_i$. For the sake of brevity, we will drop the index $i$ in the subsequent discussions.
	
Our objective is to characterize the ground-truth image $\mathbf{g}$ from the knowledge of the reduct $\mathcal{P}$. We define the reduct $\mathcal{P}$ as the subset of condition attribute sets $\mathcal{C}$.
	
We are interested in identifying the most discriminative color channel that is strongly dependent on the sky/cloud decision attribute $\mathbf{g}$. This dependence on a particular color channel is measured by the corresponding \emph{relevance} criterion of the color channels. Color channels with a high relevance value are better candidates for sky/cloud segmentation. 

In analogy to Eq.~(\ref{eq:rel-defn}), we define the relevance criterion $\gamma_j$ for each color channel $\mathbf{c}_j$, which indicates the dependence between ground truth $\mathbf{g}$ and color channels:
\begin{align}
\label{eq:rel}
\gamma_j = \frac{|\mbox{POS}(\mathbf{g})|}{|\mathcal{U}|}.
\end{align}
	
For image $\mathbf{I}_i \in {\rm I\!R}^{r \times s}$, we generate the decision table $\mathcal{L}_i \in {\rm I\!R}^{rs \times 17}$. Each row of the decision table $\mathcal{L}_i$ corresponds to a pixel. The first $16$ columns represent the color channels, and the $17^\mathrm{th}$ column corresponds to the ground truth label. We generate $16$ distinct partitions from this decision table using the knowledge of the ground truth labels. Subsequently, the corresponding lower approximations of $16$ color channels are generated using Eq.~(\ref{eq:low-upp}). In fact, this lower-approximation set is the union of all partitions (generated by the equivalence relation) that are possible members of the \emph{cloud} label. Next, we compute the relevance value $\gamma_j$ of all color channels for the image $\mathbf{I}_i$ using Eq.~(\ref{eq:rel}). 
	
We perform this for all images of dataset $\mathcal{T}$. Finally, we compute the average relevance $\overline{\gamma}_j$ of each color channel across all the images. The color channel with the maximum average relevance value $\overline{\gamma}_j$ is the best amongst all color channels under consideration.
	
\section{Experiments}
\label{sec:exp}
\subsection{Dataset}
In order to check the efficacy of our proposed color channel selection algorithm, we conduct experiments on a publicly available sky/cloud image database called HYTA~\cite{Li2011}. It consists of $32$ sky/cloud images with varying sky conditions, along with their corresponding segmentation masks (cf.\ Fig.~\ref{fig:objective}). The images were collected by sky cameras located in Beijing and Conghua, China.

\subsection{Cloud Classification Performance}
In this section, we evaluate the discriminative power of each of the $16$ color channels with respect to the cloud classification task.	
We target a single color channel for sky/cloud image segmentation, because in our previous work~\cite{ICIP1_2014,JSTARS2016} we observed that there is no significant improvement when using a combination of multiple color channels.

We follow a supervised learning approach and train a Support Vector Machine (SVM) to validate our findings. We use each of the $16$ color channels separately as candidate feature vectors and train $16$ different SVMs. The trained SVMs are then used for sky/cloud pixel classification in order to check the efficacy of respective color channels in the classification task. We randomly  segregate the HYTA dataset into 15 training images and 17  test images.

For an objective evaluation of our algorithm, we report the F-score and Accuracy. Suppose $TP$, $FP$, $TN$, and $FN$ denote the true positives, false positives, true negatives, and false negatives, respectively, in this binary classification task. Accuracy is defined as the ratio of pixels that are correctly classified, $(TP + TN)/(TP + TN + FP + FN)$. The F-score, a popular metric in a binary classification problems, is defined as the harmonic mean of Precision $= TP/(TP + FP)$ and Recall $= TP/(TP + FN)$.

\begin{figure}[htb]
\centering
\subfloat[Accuracy]{\includegraphics[width=0.5\textwidth]{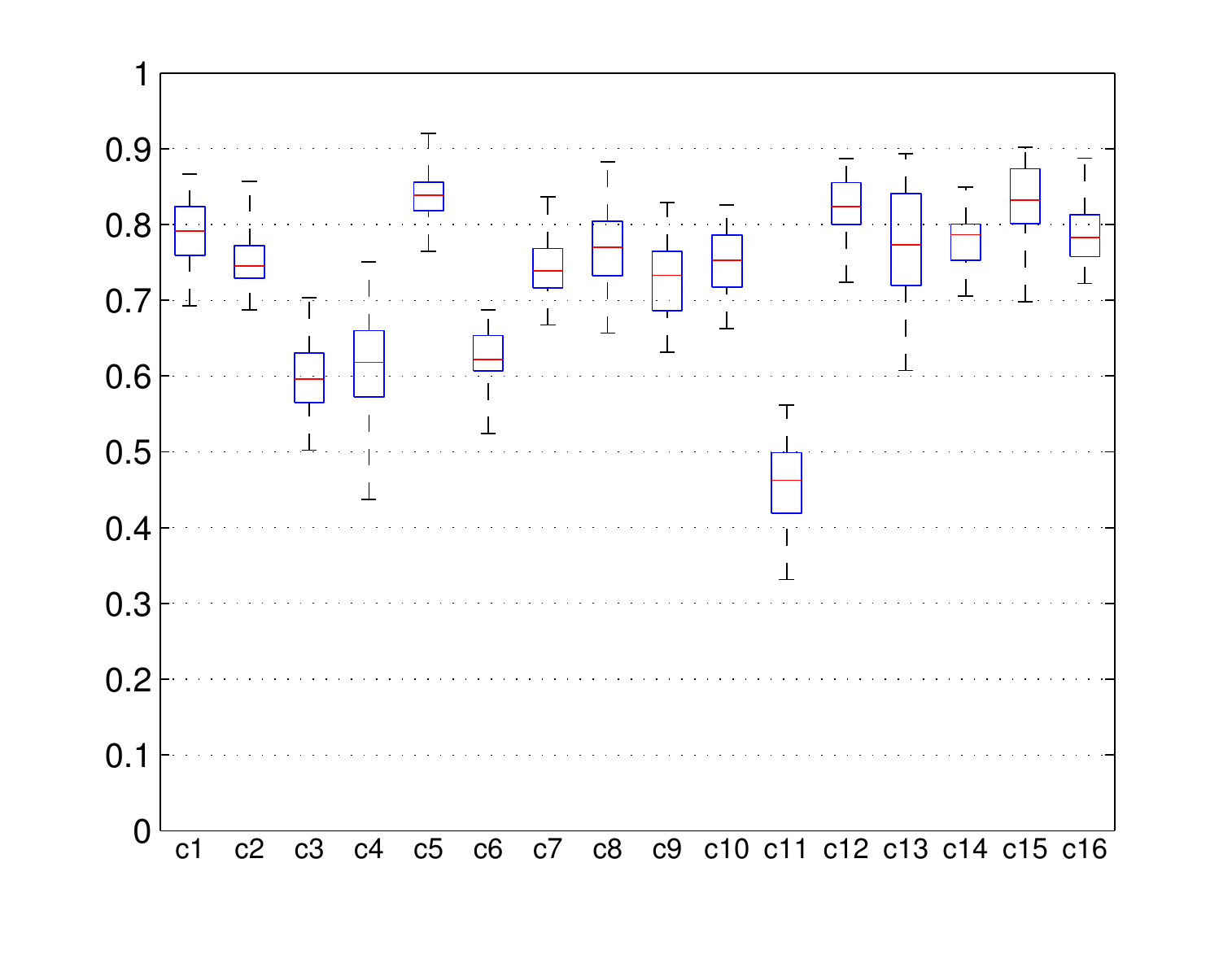}}\\
\vspace{-4mm}
\subfloat[F-score]{\includegraphics[width=0.5\textwidth]{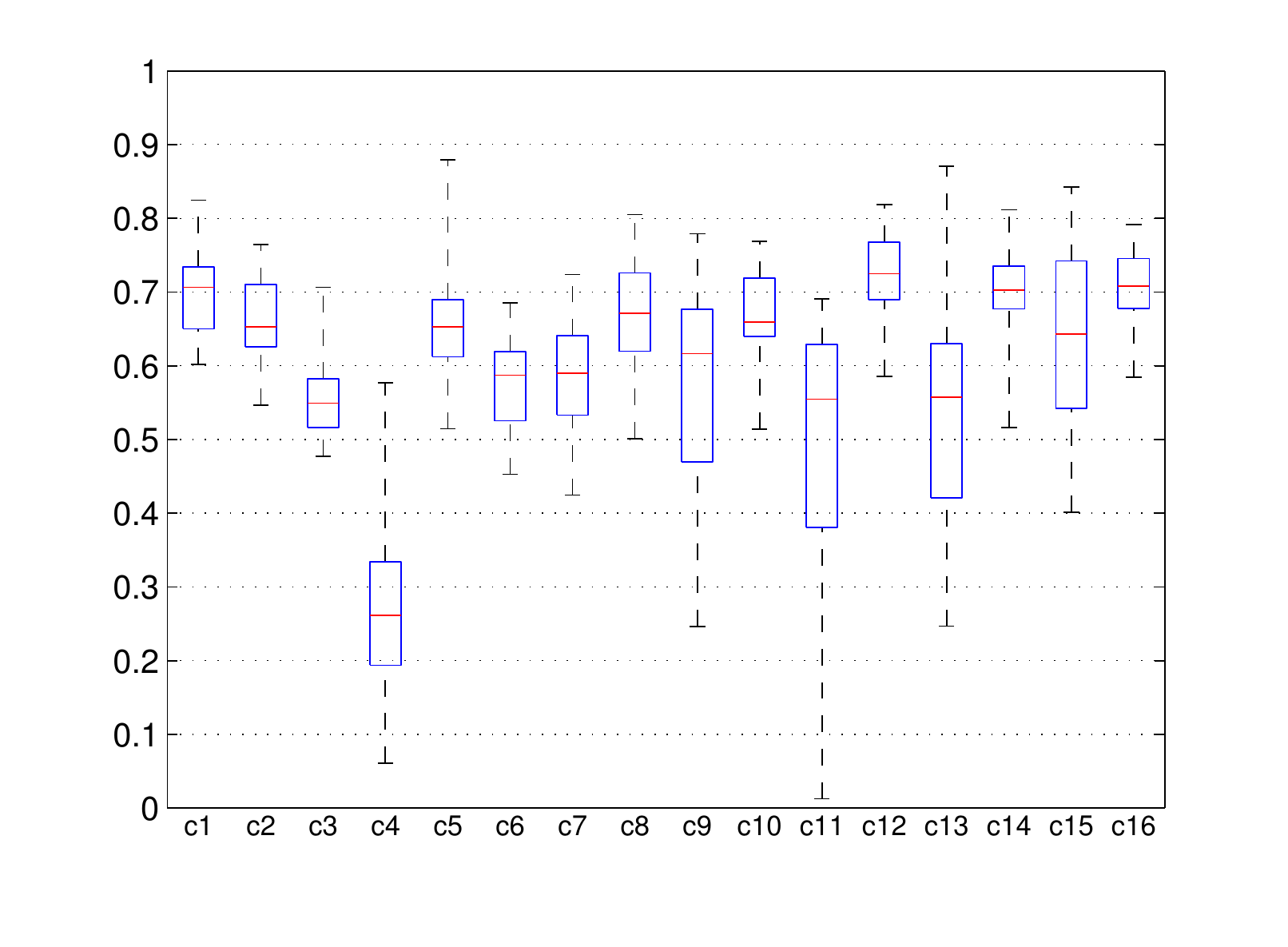}}
\caption{(a) Accuracy and (b) F-scores for $16$ color channels for all images in the HYTA database. For each box, the central red line indicates the median, the top and bottom edges correspond to $25^\textrm{th}$ and $75^\textrm{th}$ percentiles, and the whiskers represent the extreme data points.}
\label{fig:boxplots}
\end{figure}

Figure~\ref{fig:boxplots} shows the binary classification results for each of the $16$ color channels, computed over $50$ different random selections of training and test sets.

We observe from Fig.~\ref{fig:boxplots} that color channels $c_{15}$ ($\frac{B-R}{B+R}$) and $c_5$ ($S$) have relatively higher accuracy and smaller variation as compared to the other color channels. Conversely, color channels $c_{11}$ ($a^*$) and $c_4$ ($H$) score poorly in terms of accuracy. Therefore, each color channel has different discriminative power to classify sky and cloud pixels from ground-based images. This ranking of color channels in terms of their classification performance serves as the ground-truth to verify the efficacy of different color channel selection algorithms.

\subsection{Benchmarking}	
In our proposed rough set based method, we measure the dependency of each color channel and report its corresponding average relevance value $\overline{\gamma}$ across all images of HYTA. Table~\ref{tab:rel-val} summarizes the results. We observe that certain color channels viz.\ $c_{15}$ ($\frac{B-R}{B+R}$), $c_{13}$ ($\frac{R}{B}$), $c_5$ ($S$) have higher relevance scores as compared to others, making these color channels  favorable candidates for sky/cloud image segmentation.
On the other hand, color channels $c_{11}$ ($a^{*}$) and $c_4$ ($H$) have low relevance scores, indicating that these color channels contribute less to the decision attribute for (non-) cloud pixels. Therefore, the latter color channels are not conducive for sky/cloud segmentation. 

\begin{table}[htb]
\small
\centering
\begin{tabular}{ p{1mm} | p{4mm} || p{1mm} | p{4mm} || p{1mm} | p{4mm} || p{2mm} | p{4mm}|| p{2mm} | p{4mm} ||p{2mm} | p{4mm} }
\hline
$c_{1}$ & 0.70 & $c_{4}$ & 0.46 & $c_{7}$ & 0.72 & $c_{10}$ & 0.66 & $c_{13}$ & \textbf{0.84} & $c_{16}$ & 0.66\\
$c_{2}$ & 0.66 & $c_{5}$ & \textbf{0.82} & $c_{8}$ & 0.78 & $c_{11}$ & 0.33 & $c_{14}$ & 0.69 & $ $ & $ $\\
$c_{3}$ & 0.58 & $c_{6}$ & 0.58 & $c_{9}$ & 0.69 & $c_{12}$ & 0.61 & $c_{15}$ & \textbf{0.84} & $ $ & $ $\\
\hline
\end{tabular}
\caption{Average relevance value across all images of the HYTA database. The most relevant color channels are highlighted in bold.}
\label{tab:rel-val}
\end{table}

We verify this by obtaining the correlation of the individual relevance values of color channels with the sky/cloud segmentation results, and we benchmark our proposed rough set based algorithm with prior works on color channel selection~\cite{ICIP1_2014}. 

As we are interested in a binary segmentation, color channels exhibiting higher bimodal behavior are favorable. Pearson's Bimodality Index (PBI)~\cite{PBI} for all $16$ color channels is computed. A PBI value close to unity indicates highest bimodality, and values higher than $1$ indicate departure from bimodal behavior. 

Principal Component Analysis (PCA) is used to determine the color channels that capture the highest variance. We compute the absolute value of the loading factors, defined as the re-projections of the data point on the principal component axes, for the $16$ color channels, and consider the component on the first principal component axis. Those color channels with high loading factors  on the first eigenvector are considered favorable color channels.

We also compute the area under ROC curve individually for the $16$ color channels. A higher area under ROC curve for a particular color channel indicates that its classifier performance is better than the random classifier. 

Lastly, the KL-divergence of the color channels from the binary ground truth images is computed. This distance can be interpreted as the amount of dissimilarity of the color channel from the ground truth. In other words, a higher distance indicates an unfavorable color channel for cloud segmentation.
	
We check the correlation of the average accuracy scores for all the color channels with the normalized scores obtained from different approaches. Table~\ref{tab:corr_results} shows the correlation coefficients of these approaches with the average classification accuracy of the color channels. Our proposed method using rough sets achieves the highest correlation.

\begin{table}[htb]
\normalsize
\centering
\begin{tabular}{lc}
\hline
\textbf{Methods} & \textbf{Correlation} ($r$) \\
\hline 
Proposed approach & $0.84$ ($\uparrow$)\\
Bimodality & $-0.58$ ($\downarrow$)\\
Loading factors & $0.57$ ($\uparrow$)\\
ROC curve & $0.78$ ($\uparrow$) \\
Kullback-Leibler divergence & $0.12$ ($\downarrow$) \\
\hline
\end{tabular}
\caption{Correlation of cloud classification accuracy with ranking scores obtained using different methods. The \mbox{$\uparrow$ (or $\downarrow$)} indicates if \mbox{higher (or lower)} magnitude signifies better performance.}
\label{tab:corr_results}
\end{table}
 
\begin{figure*}[htb]
	\centering
	\subfloat[Relevance]{\includegraphics[height=5cm]{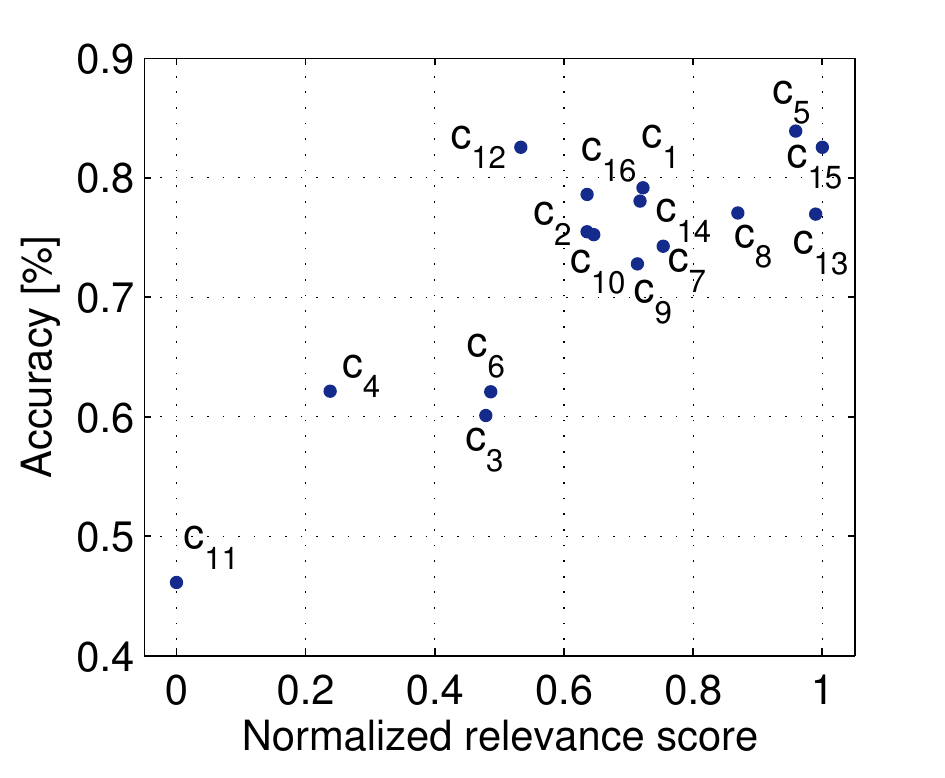}}
	\subfloat[Bimodality]{\includegraphics[height=5cm]{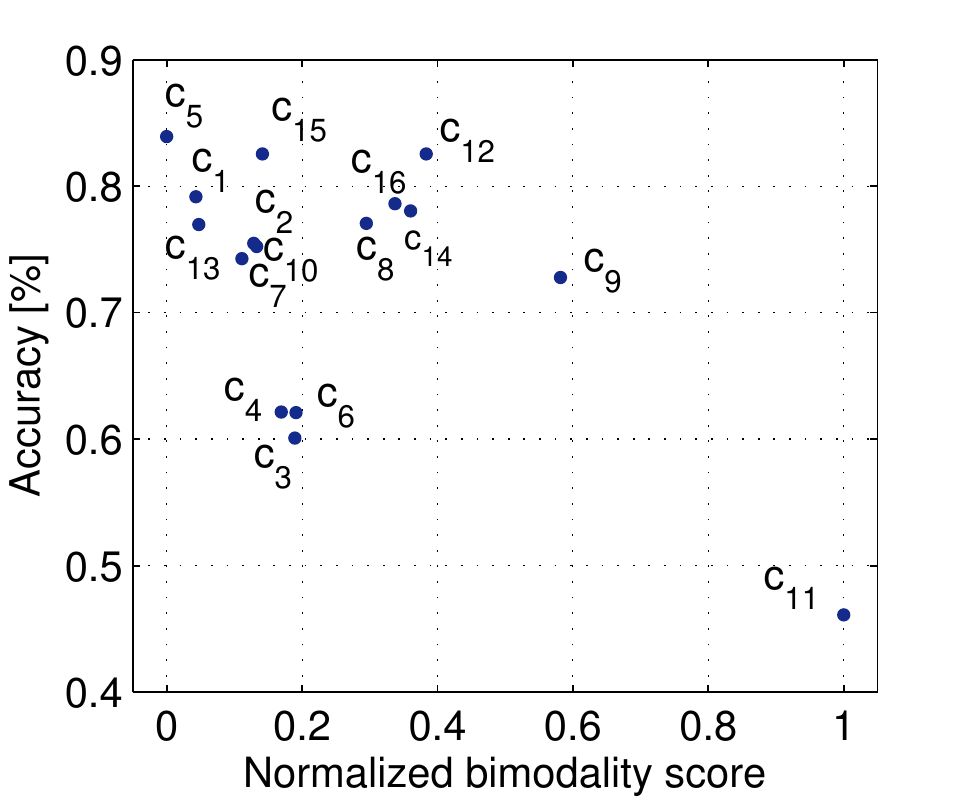}}
	\subfloat[Kullback-Leibler distance]{\includegraphics[height=5cm]{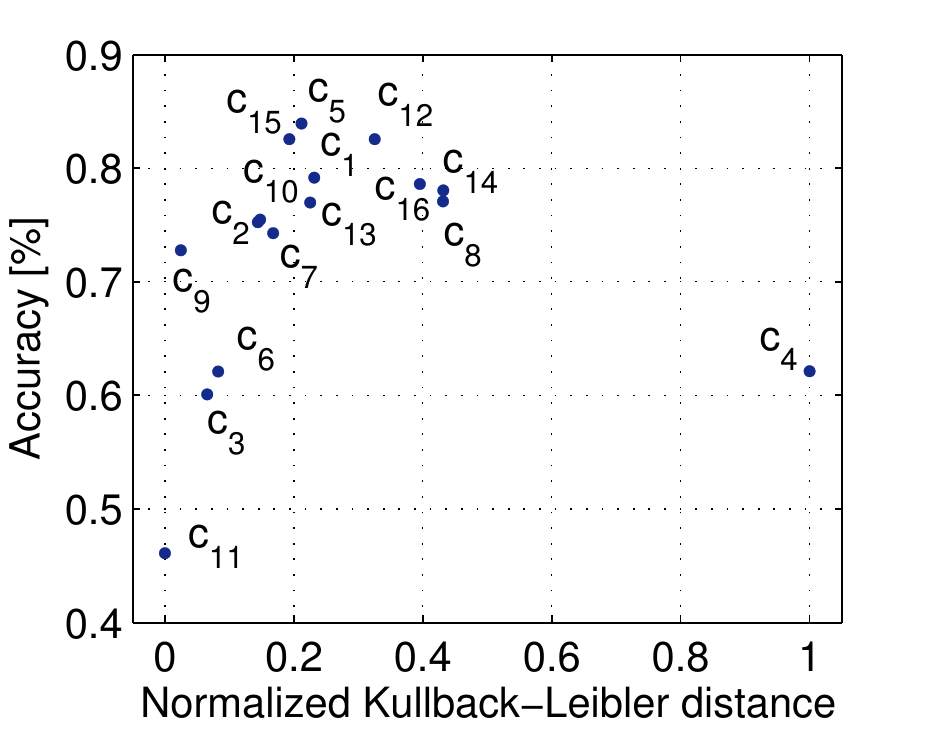}}
	\caption{Scatter plot between average accuracy and (a) relevance ($r=0.84$), (b) bimodality ($r=-0.57$), and (c) Kullback-Leibler distance ($r=0.12$) for all $16$ color channels (cf.\ Table~\ref{tab:corr_results}). Our proposed approach using relevance scores  achieves the highest correlation when ranking the color channels with respect to their segmentation performance.} 
	\label{fig:corr}
\end{figure*}

Figure~\ref{fig:corr} shows the respective scatter plots. The correlation coefficient is highest ($r=0.84$) for our proposed algorithm. From Fig.~\ref{fig:corr}(a), we can clearly see that color channels $c_5$ ($S$) and $c_{15}$ ($\frac{B-R}{B+R}$) are better candidates for sky/cloud classification. Similarly, color channels $c_{11}$  and $c_4$ with lower relevance scores are poor candidates. Figure~\ref{fig:corr}(b) also reveals that $c_5$ and $c_{15}$ are conducive color channels as their PBI values are closer to $1$. Similar results can be drawn from Fig.~\ref{fig:corr}(c), where favorable color channels $c_5$ and $c_{15}$ have comparatively lower KL-divergence values, and color channel $c_4$ has the highest KL-distance. However, the order of ranking for the other color channels is poor, leading to a low correlation value ($r=0.12$). 
	
From these results, we observe that the relevance criterion has the highest correlation with the cloud segmentation accuracy among all methods. Therefore we conclude that our proposed rough set based color channel selection algorithm is useful to rank  color channels and identify suitable ones for image segmentation. We also note that certain color channels viz.\ $c_5$ and $c_{15}$ always perform best in all three benchmarking methods. Conversely, color channels $c_4$ and $c_{11}$ rank lower than others. Visual inspection of these color channels confirms these findings (cf.\ Fig.~\ref{fig:objective}).

\section{Conclusions}
\label{sec:conclusion}
Sensing the earth's atmosphere using ground-based visible-light images is popular because of its low cost and the high temporal and spatial resolution of the captured images, as compared to traditional satellite images. In this letter, we have proposed a color channel selection algorithm based on  rough set theory. Experimental results show the efficacy of our approach in identifying favorable color channels for image segmentation. Our proposed approach outperforms other feature selection algorithms. Future work involves the extension of such rough set based approaches to other applications. 

\section*{Acknowledgment}
This work is supported by a grant from Singapore's Defence Science \& Technology Agency (DSTA).
	
\balance


\begin{thebibliography}{10}
	
	\bibitem{GRSM2016}
	S.~Dev, B.~Wen, Y.~H. Lee, and S.~Winkler,
	\newblock ``Ground-based image analysis: A tutorial on machine-learning
	techniques and applications,''
	\newblock {\em IEEE Geoscience and Remote Sensing Magazine}, vol. 4, no. 2, pp.
	79--93, June 2016.
	
	\bibitem{IGARSS2015}
	S.~Dev, F.~M. Savoy, Y.~H. Lee, and S.~Winkler,
	\newblock ``Design of low-cost, compact and weather-proof whole sky imagers for
	{High-Dynamic-Range} captures,''
	\newblock in {\em Proc. International Geoscience and Remote Sensing Symposium
		(IGARSS)}, 2015, pp. 5359--5362.
	
	\bibitem{infrared_UK}
	E.~Rumi, D.~Kerr, J.~M. Coupland, A.~P. Sandford, and M.~J. Brettle,
	\newblock ``Automated cloud classification using a ground based infra-red
	camera and texture analysis techniques,''
	\newblock in {\em Proc. SPIE Remote Sensing of Clouds and the Atmosphere
		XVIII}, 2013, vol. 8890.
	
	\bibitem{WAHRSIS}
	S.~Dev, F.~M. Savoy, Y.~H. Lee, and S.~Winkler,
	\newblock ``{WAHRSIS}: A low-cost, high-resolution whole sky imager with
	near-infrared capabilities,''
	\newblock in {\em Proc. IS\&T/SPIE Infrared Imaging Systems}, 2014, vol. 9071.
	
	\bibitem{thincloud}
	Q.~Li, W.~Lu, J.~Yang, and J.~Z. Wang,
	\newblock ``Thin cloud detection of all-sky images using {Markov} random
	fields,''
	\newblock {\em IEEE Geoscience and Remote Sensing Letters}, vol. 9, no. 3, pp.
	417--421, May 2012.
	
	\bibitem{JSTARS2016}
	S.~Dev, Y.~H. Lee, and S.~Winkler,
	\newblock ``Color-based segmentation of sky/cloud images from ground-based
	cameras,''
	\newblock {\em IEEE Journal of Selected Topics in Applied Earth Observations
		and Remote Sensing}, vol. PP, no. 99, pp. 1--12, 2016.
	
	\bibitem{ICIP2015b}
	S.~Dev, Y.~H. Lee, and S.~Winkler,
	\newblock ``Categorization of cloud image patches using an improved
	texton-based approach,''
	\newblock in {\em Proc. International Conference on Image Processing (ICIP)},
	2015, pp. 422--426.
	
	\bibitem{IGARSS2015b}
	F.~M. Savoy, J.~Lemaitre, S.~Dev, Y.~H. Lee, and S.~Winkler,
	\newblock ``Cloud base height estimation using high-resolution whole sky
	imagers,''
	\newblock in {\em Proc. International Geoscience and Remote Sensing Symposium
		(IGARSS)}, 2015, pp. 1622--1625.
	
	\bibitem{Kreuter2009}
	A.~Kreuter, M.~Zangerl, M.~Schwarzmann, and M.~Blumthaler,
	\newblock ``All-sky imaging: A simple, versatile system for atmospheric
	research,''
	\newblock {\em Applied Optics}, vol. 48, no. 6, pp. 1091--1097, Feb. 2009.
	
	\bibitem{Li2011}
	Q.~Li, W.~Lu, and J.~Yang,
	\newblock ``A hybrid thresholding algorithm for cloud detection on ground-based
	color images,''
	\newblock {\em Journal of Atmospheric and Oceanic Technology}, vol. 28, no. 10,
	pp. 1286--1296, Oct. 2011.
	
	\bibitem{Calbo2008}
	J.~Calb\'{o} and J.~Sabburg,
	\newblock ``Feature extraction from whole-sky ground-based images for
	cloud-type recognition,''
	\newblock {\em Journal of Atmospheric and Oceanic Technology}, vol. 25, no. 1,
	pp. 3--14, Jan. 2008.
	
	\bibitem{Heinle2010}
	A.~Heinle, A.~Macke, and A.~Srivastav,
	\newblock ``Automatic cloud classification of whole sky images,''
	\newblock {\em Atmospheric Measurement Techniques}, vol. 3, no. 3, pp.
	557--567, 2010.
	
	\bibitem{LiuSP2015}
	S.~Liu, L.~Zhang, Z.~Zhang, C.~Wang, and B.~Xiao,
	\newblock ``Automatic cloud detection for all-sky images using superpixel
	segmentation,''
	\newblock {\em IEEE Geoscience and Remote Sensing Letters}, vol. 12, no. 2, pp.
	354--358, Feb. 2015.
	
	\bibitem{Souza}
	M.~P. Souza-Echer, E.~B. Pereira, L.~S. Bins, and M.~A.~R. Andrade,
	\newblock ``A simple method for the assessment of the cloud cover state in
	high-latitude regions by a ground-based digital camera,''
	\newblock {\em Journal of Atmospheric and Oceanic Technology}, vol. 23, no. 3,
	pp. 437--447, March 2006.
	
	\bibitem{ICIP1_2014}
	S.~Dev, Y.~H. Lee, and S.~Winkler,
	\newblock ``Systematic study of color spaces and components for the
	segmentation of sky/cloud images,''
	\newblock in {\em Proc. International Conference on Image Processing (ICIP)},
	2014, pp. 5102--5106.
	
	\bibitem{fs-ROC}
	A.~J. Serrano, E.~Soria, J.D. Martin, R.~Magdalena, and J.~Gomez,
	\newblock ``Feature selection using {ROC} curves on classification problems,''
	\newblock in {\em Proc. International Joint Conference on Neural Networks
		(IJCNN)}, 2010.
	
	\bibitem{Kerekes08}
	J.~Kerekes,
	\newblock ``Receiver operating characteristic curve confidence intervals and
	regions,''
	\newblock {\em IEEE Geoscience and Remote Sensing Letters}, vol. 5, no. 2, pp.
	251--255, April 2008.
	
	\bibitem{InfoTheory_TGRS}
	A.~Mart{\'i}nez-Us{\'o}, F.~Pla, J.~M. Sotoca, and P.~Garc{\'i}a-Sevilla,
	\newblock ``Clustering-based hyperspectral band selection using information
	measures,''
	\newblock {\em IEEE Transactions on Geoscience and Remote Sensing}, vol. 45,
	no. 12, pp. 4158--4171, Dec. 2007.
	
	\bibitem{Pawlak92}
	Z.~Pawlak,
	\newblock {\em Rough Sets: Theoretical Aspects of Reasoning About Data},
	\newblock Kluwer Academic Publishers, 1992.
	
	\bibitem{rs-TGRS}
	S.~Patra, P.~Modi, and L.~Bruzzone,
	\newblock ``Hyperspectral band selection based on rough set,''
	\newblock {\em IEEE Transactions on Geoscience and Remote Sensing}, vol. 53,
	no. 10, pp. 5495--5503, Oct. 2015.
	
	\bibitem{PBI}
	T.~R. Knapp,
	\newblock ``Bimodality revisited,''
	\newblock {\em Journal of Modern Applied Statistical Methods}, vol. 6, no. 1,
	2007.
	
\end{thebibliography}
\end{document}